\documentclass[a4paper,fleqn, review]{cas-dc}

\usepackage[square,sort,comma,numbers]{natbib}
\setcitestyle{square}
  \usepackage{todonotes}
\DeclareUnicodeCharacter{2061}{}
\usepackage{tikz}
\usetikzlibrary{shapes.geometric, arrows}
\usepackage{subfig}
\usepackage{balance}
\usepackage{algpseudocode}
\usepackage{algorithm}
\usepackage{bm}
\algdef{SE}[VARIABLES]{Variables}{EndVariables}
   {\algorithmicvariables}
   {\algorithmicend\ \algorithmicvariables}
\algnewcommand{\algorithmicvariables}{\textbf{Variables}} 

\usepackage{caption}
\def\tsc#1{\csdef{#1}{\textsc{\lowercase{#1}}\xspace}}
\tsc{WGM}
\tsc{QE}

\begin{document}
\let\WriteBookmarks\relax
\def\floatpagepagefraction{1}
\def\textpagefraction{.001}

\shorttitle{A Bayesian Active Machine Learning Approach to Comparative Judgement}


\title [mode = title]{A Bayesian Active Learning Approach to Comparative Judgement}   



%

\ifdefined\DOUBLEBLIND
    \shortauthors{Anonymous \textit{et al.}}    
    \author[]{-- Author Names Removed for Peer-Review --}
\else
    \shortauthors{Gray \textit{et al.}}

\author[1]{Andy Gray}[type=editor,
    orcid=0000-0002-1150-2052,
    twitter=codingWithAndy,
]

\cormark[1]


\ead{445348@swansea.ac.uk}



\affiliation[1]{organization={Swansea University},
            city={Swansea},
            country={United Kingdom}}

\author[1]{Alma Rahat}[
    orcid=0000-0002-5023-1371,
    twitter=AlmaRahat,
    ]


\ead{a.a.m.rahat@swansea.ac.uk}






\author[1]{Tom Crick}[
    orcid=0000-0001-5196-9389,
    twitter=ProfTomCrick,]


\ead{thomas.crick@swansea.ac.uk}

\author[2]{Stephen Lindsay}[orcid=0000-0001-6063-3676]
\affiliation[2]{organization={Glasgow University},
            city={Glasgow},
            country={United Kingdom}}

\ead{stephen.lindsay@glasgow.ac.uk}




\fi

\begin{abstract}
    Assessment is a crucial part of education. Traditional marking is a source of inconsistencies and unconscious bias placing a high cognitive load on the assessors. One approach to address these issues is comparative judgement (CJ). In CJ, the assessor is presented with a pair of items of work, and asked to select the better one. Following a series of comparisons, a rank for any item may be derived using a ranking model, for example, the Bradley-Terry model, based on the pairwise comparisons. While CJ is considered to be a reliable method for conducting marking, there are concerns surrounding its transparency, and the ideal number of pairwise comparisons to generate a reliable estimation of the rank order is not known. Additionally, there have been attempts to generate a method of selecting pairs that should be compared next in an informative manner, but some existing methods are known to have created their own bias within results inflating the reliability metric used within the process. As a consequence, a random selection approach is usually deployed. 

    In this paper, we propose a novel Bayesian approach to CJ (which we call BCJ) for determining the ranks of a range of items under scrutiny alongside a new way to select the pairs to present to the marker(s) using active learning, addressing the key shortcomings of traditional CJ. Furthermore, we demonstrate how the entire approach may provide transparency by providing the user insights into how it is making its decisions and, at the same time, being more efficient. Results from our synthetic experiments confirm that the proposed BCJ combined with entropy-driven active learning pair-selection method is superior (i.e. always equal to or significantly better) than other alternatives, for example, the traditional CJ method with differing selection methods such as uniformly random, or the popular no repeating pairs where pairs are selected in a round-robin fashion. We also find that the more comparisons that are conducted, the more accurate BCJ becomes, which solves the issue the current method has of the model deteriorating if too many comparisons are performed. As our approach can generate the complete predicted rank distribution for an item, we also show how this can be utilised in probabilistically devising a predicted grade, guided by the choice of the assessor.
\end{abstract}





\maketitle

\section{Introduction}
    \label{sec:intro}
        Inconsistency in teachers predicting student grades is widespread. In schools and collages across the UK in 2019, only $21\%$ of students obtained the grades that were predicted by their teachers~\cite{bbc_a-levels}. A study in 2011 found that $42$-$44\%$ of teacher grades over-predicted by at least one grade, and $7$-$11\%$ under-predicted~\cite{everett2011investigating}. The immediate impact of the COVID-19 pandemic across educational settings and contexts globally was profound~\cite{watermeyer-et-al:he2020,crick-et-al:ukicer2020,marchant-et-al:plosone2021,crick-et-al:educon2021,wg1final:iticsewgr2021,lowthian-et-al:jrsm2023} and its long-term impact has still not fully manifested~\cite{watermeyer-et-al:bjse2021,shankar-et-al:ies2021,mcgaughey-et-al:herd2022,hardman-et-al:sajhe2022}; we will likely continue to experience a ``new normal'' for education over the coming period~\cite{itnowdigedu:2021,phillips-et-al:educon2021,irons+crick:cyberbook2022}, and especially for educational assessment~\cite{watermeyer-et-al:ijad2022,wg1final:iticsewgr2021,crick-et-al:iticse2022,ward-et-al:jime2023}. During the COVID-19 pandemic, student grades were given based on teachers' assessments in England and Wales (two of the four nations of the UK, with separate education systems), resulting in record-high grades for GCSE and A-level students.  However, with the announcement of the 2022 A-level results, $80,000$ fewer students obtained grades of $A^*$ or $A$ grade compared to 2021, a fall from $19.1\%$ getting $A^*$s in 2021 compared to $13.5\%$ in 2022 -- ultimately bringing grades back in line with pre-pandemic results in 2019~\cite{guardian_a-level}. However, it is increasingly clear that there is subjectivity, bias and inequity when it comes to making an overall judgement on a pupil's performance~\cite{lsecovidblog:2022}; indeed, asking fundamental questions such as: is assessment fair?~\cite{assessfair:2020}. 

        There is an extensive corpus of work that focuses on using intelligent and/or data-driven approaches in a variety of educational settings and contexts~\cite{luckin-et-al:2016,namoun2020predicting,rastrollo2020analyzing,dwivedi-et-al:ijim2019,shafiq2022student}; in particular, for predicting student performance and retention we have seen broad application of data mining and learning analytics~\cite{elbadrawy2016predicting,smartlearn:2022}, as well as machine learning, collaborative filtering, recommender systems, and artificial neural networks~\cite{iqbaletal2017mlcasestudy,mlcomp:2019,yousafzai-et-al:2020eit}. However, there exists a number of increasingly complex and interconnected social, ethical, legal and digital/data rights issues with these varied approaches~\cite{slade+prinsloo:2013,williamson2020datafication,akgun+greenhow:2022}, especially with a pre- and post-pandemic critical analysis~\cite{williamson2020pandemic}. It is also potentially problematic, in a educational policy context, to be perceived to be disempowering educators and undermining their expertise in supporting learning and progression via formative and summative assessment approaches.
        
        Prospect theory shows that humans are better at identifying relative, rather than absolute quality~\cite{kahneman2013prospect}. In the educational assessment context, this has been recognised~\cite{benton2018comparative}, and comparative judgement (CJ) has been proposed as an alternative to traditional marking~\cite{pollitt1996raters}. In CJ, an assessor is presented with a pair of items of work, and they only make a decision on which one is of higher quality instead of assigning an absolute mark. The process is repeated some predefined number of iterations, potentially re-evaluating already evaluated pairs. A ranked order of items is then derived from these pairwise comparisons using a model of CJ, for example, the Bradley-Terry model~\cite{hunter2004mm}, which was inspired from Thurstone's mathematical definition of generating ranks from comparisons~\cite{thurstone1927law}. In this way,
        we are able to extract an accurate ranked order from only a series of relative comparisons. 
        In addition, an important benefit of CJ is that the cognitive load placed on the teachers while marking is also reduced~\cite{coenen2018information}.  
        
        Nonetheless, one of the key drawbacks of CJ is that, irrespective of specific approaches, it can take numerous interactions (i.e. the number of pairs to be assessed) and significant time for completing marking, on top of the time required to collate grades, award students’ scores, and then provide feedback. Alternative methods of CJ e.g. adaptive comparative judgement (ACJ) are designed to reduce interactions without loss of accuracy, but have been found to include other bias through their ``adaptive nature''~\cite{bramley2015investigating}. Hence, the pure form is still the desired version. This means, while CJ has its benefits, a method that reduces the number of interactions and overall time it takes to mark is still an important open research problem. 

        Furthermore, Ofqual, the official governmental body that regulates qualifications, exams and tests in England, has also pointed out that CJ's paired comparison rank order starts to deteriorate, and the whole model fit starts to collapse unless it is precisely known what the minimum number of judgements needed is in advance, and with confidence which is not known~\cite{ofqual2017}. Additionally, Ofqual also believes that CJ has issues with being less transparent in how it makes and presents its findings~\cite{ofqual2017}. 

        We thus propose a novel Bayesian approach towards CJ -- which we name BCJ -- addressing the key weaknesses of traditional CJ. Our primary aims in developing BCJ were reducing interactions and providing greater insight into the ranking decision process. The main contributions of this paper are as follows: 
        \begin{itemize}
                \item We derived an analytical expression to compute the entire {\emph{predictive rank distribution}} for any item that is being assessed with densities over pairwise preferences.
                \item We illustrate how each of these pairwise preference densities, and as a consequence the overall rank distributions for an item, can be updated via Bayesian methodology, as we collect more data on pairwise comparisons. 
                \item We propose a novel active learning (AL) approach, based on predictive entropy of the pairwise preference densities i.e. a measure of the average uncertainty about the outcome of the contest, to select the next pair that should be assessed.
                \item We propose a probabilistic approach based on the predictive rank distributions to assign a grade to each item controlled by the assessor. 
        	\item For the first time, we demonstrate through repeated experiments on a range of synthetic problems that the proposed BCJ AL framework with entropy-based selection method is statistically the best (or equivalent to the best) for all configurations. 
        \end{itemize}
        
        The rest of the paper is structured as follows; in Section~\ref{sec:related_work}, we present the related work of the study and some background; Section~\ref{sec:gen_ranks} outlines how the main algorithms work to rank students' work. We will explain the three methods used for selecting the next pairs to be compared in Section~\ref{sec:pair_method}; we present our results and discussions in Section~\ref{sec:experiment_discussion}, with general conclusions and future work articulated in Section~\ref{sec:conclusion}.

\section{Related Work in Education}
    \label{sec:related_work}
        
        CJ is a technique used to derive ranks from pair-wise comparisons. The concept of CJ is used in academic settings to allow teachers to compare two pieces of work and select which is better against selected criteria. After each comparison, another pair is selected. This is repeated until enough pairs have been compared to generate a ranking of the work marked. We detail a typical CJ process in Algorithm~\ref{alg:opt}.

        \begin{algorithm} [h!]
  \caption{Standard comparative judgement procedure.}
  \label{alg:opt}
  \textbf{Inputs.}
  \begin{algorithmic}[]
    \State $N:$ Number of items.
    \State $K:$ Multiplier for computing the budget for the number of pairs to be assessed.
    \State $I:$ Set of items.
  \end{algorithmic}
  \bigskip
  \textbf{Steps.}
  \begin{algorithmic}[1]
    \State $B \gets N \times K$ \Comment{\small{Compute the budget.}}
 \State $G \gets \langle \rangle$ \Comment{\small{Initialise list of selected pairs.}}
 \State $W \gets \langle \rangle$ \Comment{\small{Initialise list of  winners.}}
 \State $\mathbf{r} \gets \left(\frac{N}{2}, \dots, \frac{N}{2}\right)^\top ~|~ 
 \lvert \mathbf{r} \rvert = N$  \par
        \hskip\algorithmicindent\Comment{\small{Initialise rank vector with mean rank for all items.}}
	\For{$b = 1 \rightarrow B$}
            \State $(i, j) \gets \text{SelectPair}(I)$ \label{alg:sel_pair}
            \Comment{\small{Pick a pair of items.}} \label{alg:pick_pair}
            \State $G \gets G \oplus \langle (i,j) \rangle$ \Comment{\small{Append the latest pair.}}
            \State$ w \gets \text{DetermineWinner}(i, j) $\Comment{\small{Pick a pair of items.}}
            \State $W \gets W \oplus \langle w \rangle$ \Comment{\small{Append the latest winner.}}
            \State $\mathbf{r} \gets \text{GenerateRank}(G, W) $ \Comment{\small{Update rank vector.}} \label{alg:gen_rank}
	\EndFor
	\State \Return $\mathbf{r}$
  \end{algorithmic}
\end{algorithm}
    
        An important benefit to CJ within an academic setting is reducing the teacher's cognitive load~\cite{chen-et-al:2023}, as comparing two pieces of work is faster than marking each individual piece of work, while also insisting the teacher is being non-biased towards a student and consistent~\cite{sadler:1989}. This is difficult to achieve~\cite{bramleypaired:2007}, and CJ helps, to an extent, address this challenge; for further discussion of this, we refer to the following literature where 
        the teachers can be referred to as the judges~\cite{,benton2018comparative, bartholomew2019using, christodoulou2017making}.

        CJ is based on a technique originally proposed by Thurstone in 1927, known as `the law of comparative judgement'~\cite{thurstone1927law}. Thurstone discovered that humans are better at comparing things to each other rather than making judgements in isolation, for example, judging if a piece of fruit is bigger than another without having the other fruits to compare against at the point of judgement. Therefore, he proposed making many pair-wise comparisons until a rank order has been created~\cite{thurstone1927law, benton2018comparative, bartholomew2019using}. Pollitt \textit{et al.} introduced and popularised it within an education setting~\cite{pollitt1996raters, pollitt2004let}.

        Typically, the efficacy of a CJ method is measured using the Scale Separation Reliability (SSR)~\cite{bramley2015investigating, NMM_accuracy, pollitt2012comparative}. SSR is defined as the ratio between the variance of the true score to the variance of estimated scores from observations; interested readers should refer to the work of Verhabert \textit{et al.} \cite{verhavert2018scale} for a detailed discourse on SSR. The relative uncertainty estimation through SSR is highly dependent on the underlying CJ model (e.g. BTM) and its own estimated uncertainty, which is typically not presented to the users of the system in an intuitive way. SSR might not even be calculable, as it requires the knowledge of variance of true scores, which is unavailable in most practical cases. 
        
        An important consideration in CJ is the stopping criterion. To the best of our knowledge, there seems to be no natural and meaningful performance metric that would allow a clear indication on when to stop. Because of this, CJ is usually conducted on a fixed budget giving the number of pairs that must be compared before finalising the rank order, for example, at least $10$ judgements per script~\cite{wheadon2020comparative}.

        A growing body of evidence supports using CJ as a reliable alternative for assessing open-ended and subjective tasks. The judgements recorded by teachers, more generally termed \textit{raters} or \textit{judges}, are fed into a BTM (see Section~\ref{subsec:btm} for more details on BTM) to produce scores that represent the underlying quality of the scripts~\cite{bradley1952rank, luce1959individual}. These scores have the appealing property of being equivalent across comparisons~\cite{andrich1978rating}.  

        A key justification for using CJ within the educational assessment process is that the rank orders it produces tend to have high levels of reliability. For example, in 16 CJ exercises conducted between 1998 and 2015, the 
        correlation coefficient scores were between 0.73 to 0.99 when compared with rubric based grades \cite{steedle2016evaluating}. With a correlation coefficient of $1.0$ representing perfect agreement, a score of $0.70$ or above is typically considered to be high enough to proclaim strong agreement~\cite{hinkle2003applied}. 

        Alternative methods of CJ, specifically differing on how the pairs to evaluate next are selected or allocated to assessors, have been introduced, like ACJ. ACJ is a version that aims to be adaptive based on the current state of the marking between the judges. The adaptive nature of ACJ is based on an algorithm embedded within the approach, which pairs similarly ranked items as the judge progresses in the comparative judgement process, a method aimed at expediting the process of achieving an acceptable level of reliability~\cite{bartholomew2019using}. Pollitt first proposed ACJ in 2011, a system created in partnership with TAG assessments~\cite{pollitt2012comparative}. Later, the system was further developed by RM Compare~\cite{jones_davies_2022}. 
        
        A significant flaw in the ACJ approach was that its adaptive nature generated its own bias by having similarly ranked pieces of students' work being compared against themselves more often, and thus the correlation between true reliability and SSR (due to ACJ) has been shown to be low in some experiments~\cite{bramley2015investigating}. Further, the process usually takes longer than traditional marking~\cite{benton2018comparative, bramley2015investigating}. Therefore, it is suggested that having random pairings is just as effective as the ACJ approach. As a result, the CJ community has reverted, to a degree, back to random pairings and removed the adaptive nature of the CJ process~\cite{wheadon2020comparative, jones_davies_2022}.

        Additionally, claims have been made that advocates of CJ have not compiled a compelling case to support two of their central claims that humans are better at comparative than absolute judgments and that CJ is necessarily valid because it aggregates judgments made by experts in a naturalistic way~\cite{kelly2022critiquing}. Nonetheless, there are experiments that provide clear evidence of human efficiency in CJ in general~\cite{kahneman2013prospect}, and the practical consistency of CJ for marking~\cite{steedle2016evaluating}. However, there is a lack of clarity in how the decisions are being made, and we note this as one of the key criticisms of CJ, alongside the lack of estimations of uncertainty in estimations, despite the practical strengths. Our investigation in BCJ was primarily driven by these criticisms with an aim to improve the state-of-the-art of CJ.
        
        In the following section, we will first focus on describing the generation of ranks from paired comparisons (in line~\ref{alg:gen_rank} of Algorithm~\ref{alg:opt}), as the pair selection method (in line~\ref{alg:sel_pair} of Algorithm~\ref{alg:opt}) we propose depends on the concepts required for rank generation.  

    \section{Generating Ranks From a List of Paired Comparisons}
        \label{sec:gen_ranks}
        Currently, the most popular method of ranking paired comparisons is through the use of BTM. Therefore, in this section, we will first explain how the BTM system works, and then provide a description of our proposed Bayesian approach.
        
        \subsection{Classical Approach: Bradley-Terry Model}
            \label{subsec:btm}
            Bradley and Terry proposed BTM in their seminal paper on the topic~\cite{benton2018comparative, bisson2016measuring, marshall2020assessment, pollitt2012comparative, gray2022using}. Traditionally, this has been adopted as the driving algorithm for CJ. The technique is an iterative minorisation-maximisation (MM) method~\cite{hunter2004mm} for estimating the maximum likelihood of the expected preference score $\gamma_i$ for the $ith$ student’s item of work, given the observed data. With the expected preferences, we can then use this to arrange the items of work and then generate a rank where a higher value represents a better 
            quality of work. We present a mathematical description of the model below, broadly following Hunter's work~\cite{hunter2004mm}. 
            
            Consider the set of $N$ items, $I = \{1, \dots, N\}$ with each element $i$ representing the identifier of the relevant item. The expected performance vector is $\bm{\gamma} = (\gamma_1, \dots, \gamma_N)^\top$, where $\gamma_i$ is a positive parameter representing the overall score for the $i$th item. For example, in a typical marking context, we can assume that an individual's mark vary between $0$ and $100$, i.e. $\gamma_i \in [0, 100]$; however, this assumption is not essential for the scheme to work, and thus can be safely ignored. Now, the probability that the $i$th item is of higher quality compared to the $j$th item is given by:
            \begin{equation}
            \label{eq:btm_prob}
                P(i \succ j) = \frac{\gamma_i}{\gamma_i + \gamma_j}.
            \end{equation}

            Using the key assumption that the outcomes of different pairings are independent, the log-likelihood for the performance vector $\bm{\gamma}$ is given by:
            \begin{equation}
                L(\bm{\gamma}) = \sum_{i = 1}^N \sum_{j = 1}^N \left[ \omega_{[i, j]} \ln (⁡\gamma_i) - \omega_{[i, j]} ln( \gamma_i + \gamma_j ) \right],
                \label{eq:btm_liklihood}
            \end{equation}
            where $\omega_{[i,j]}$ is the number of times item $i$ was preferred over item $j$. It should be noted that typically BTM ignores any notion of ties, and raters are forced to make a decision on the winner.
            
            The minorisation-maximisation (MM) algorithm proposed by Hunter \cite{hunter2004mm} iteratively updates each $\gamma_i$ such that the log-likelihood in \eqref{eq:btm_liklihood} is maximised. The iterative update formula for $k$th iteration is \cite{gescheider2013psychophysics}: 
    
                \begin{equation}
                    \gamma_i^{k+1} = \Omega_i \sum_{j|j \neq i} \frac{\omega_{[i,j]} + \omega_{[j,i]}}{( \gamma_i^k + \gamma_j^k)}
                    \label{eq:5}
                \end{equation}
            
            Where, $\Omega_i = \sum_j \omega_{[i,j]}$  is the number of times the $i$th item was preferred. At each iteration, we are further required to normalise the $\gamma_i$s to ensure that the 
             sum of the elements of the performance vector
            equals $1$, i.e.
    
                \begin{equation}
                    \gamma_i^{k+1} \leftarrow \frac{\gamma^{k+1}_i}{\sum_j \gamma_j^{k+1}}.
                    \label{eq:7}
                \end{equation}
            
            Under certain assumptions, the iterative process will converge to the optimal $\bm{\gamma}$~\cite{hunter2004mm}. In this work, at the final stage, for ease of presentation and assuming $\gamma_i \in [0, 100]$, we multiply $\gamma_i$ by 100. We can then extract the rank of the $i$th item as follows (using $1$-based counting):
            \begin{equation}
            \label{eq:btm_r}
               r_{i\in I} = (N + 1) - \text{argsort}(\bm{\gamma}).
            \end{equation}

            The process in Equation~\eqref{eq:btm_r} can be repeated to generate the complete rank vector in line~\ref{alg:gen_rank} of Algorithm~\ref{alg:opt}.

        \subsection{Proposed Bayesian Approach}
            While the current CJ based on BTM works well, a core weakness is that it produces a point estimate of performances through maximising the likelihood in \eqref{eq:btm_liklihood} without estimating the epistemic uncertainty in ranks due to the paucity of data. One way to estimate the uncertainty (that is not commonly used in education context) is to use a Bayesian statistical approach; interested readers should refer to \cite{van2021bayesian} for a concise and recent overview, and to \cite{mcelreath2020statistical} for a complete and accessible discourse, of the topic.
            
            Typically, the application of a Bayesian approach to CJ has entailed using \textit{prior distributions} over the performance vector $\bm{\gamma}$ (and other parameters of the likelihood function) alongside the observed data to identify a posterior distribution over $\bm{\gamma}$ using Bayes' theorem, and produces similar results to standard CJ in terms of identifying the ranking~\cite{pritikin2020exploratory, wainer2022bayesian, tsukida2011analyze, maeyer2021bayesian}.
            However, there are important barriers that make it challenging to adopt for real-world deployment. Two key issues are:
            \begin{description}
                \item[Computation Time.]  \textit{Inferring} the posterior distribution require computationally expensive sampling based approaches (e.g. Markov Chain Monte Carlo in~\cite{wainer2022bayesian}), as analytical solution to computing the posterior is usually not available in this context. This is a major issue in using this approach for practical implementations: we want to be able to indicate the ranks to the assessors quickly, possibly after each pairwise comparison, without a significant delay (e.g. several minutes). 
                \item[Modelling Performance Instead of Pairwise Preference.] In a ranking exercise, we are generally interested in identifying the ranks of the items, and the observed data is from pairwise comparisons. However, in standard CJ, including the typical Bayesian approach, the performances are modelled instead of pairwise preference; the latter is usually treated as an outcome of a latent function and thus only reflected in derived ranks from the expected (or average) performances. As a result, while it is possible to extract uncertainty estimates over the preferences or the ranks (with the aforementioned computational expense), they are never communicated or used to provide insights to the assessors. Subsequently, an opportunity to utilise the uncertainty in preference to drive the collection of new pairwise comparisons is missed. Furthermore, the performance scores yielding from these modelling do not have a direct scalar relationship to the scores of the assessment designed by the assessor. So, it is difficult to interpret these scores easily. 
            \end{description}

            Addressing these primary issues, we propose to adopt a Bayesian approach where we focus on \textit{modelling the pairwise preferences}. We expect this approach will allow us to capture most information due to the direct relationship between pairwise preference and data from pairwise comparisons. The posterior allows us to identify the predictive density over the ranks of the items. Moreover, the uncertainty estimations in preferences helps us drive the selection of the next pair to compare in an active learning manner. We discuss the selection method in Section~\ref{sec:al_ent}.

            \subsubsection{Pairwise Preference Model}

\begin{figure}[t!]
                \centering
                \includegraphics[width=0.9\columnwidth, trim={5mm 5mm 5mm 5mm}, clip=true]{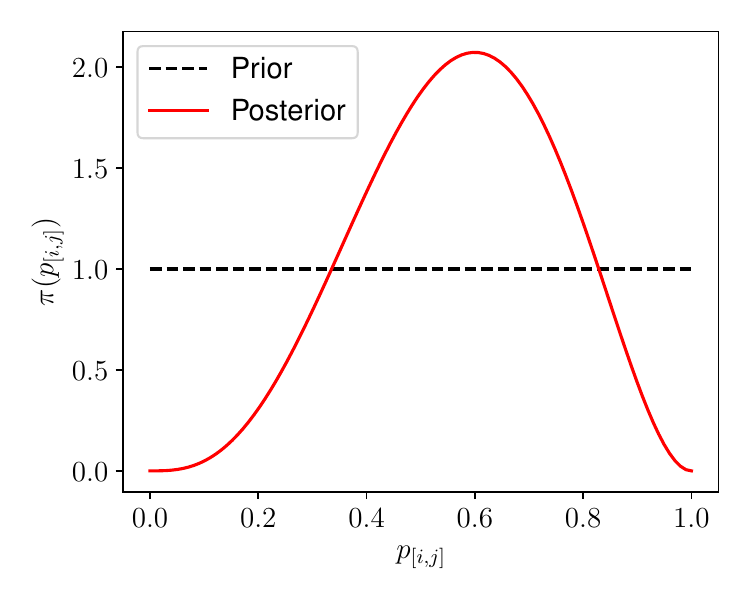}
                \caption{A toy example of Bayesian updating of PDF over preference between $i$th and $j$th items. Initially, with uniform prior (shown with a black dashed line), none is preferred. Then, with three wins ($\alpha=1+3=4$) and two losses ($\beta = 1+5-3=3$) for $i$ after five comparisons, the PDF (depicted with a red line) starts to skew in favour of $i$ (i.e. towards $1$). The more data we have, the narrower the PDF will become, i.e. the uncertainty would reduce.}
                \label{fig:ex-prior-post}
            \end{figure}

            Let the outcome of a paired comparison between the $i$th and $j$th item be binary, i.e. $x=0$, or $x=1$, with $x=1$ representing a preference for $i$, and \textit{vice versa}. Now, considering the data $\mathbf{x} = (x_1, \dots, x_n)^\top$ as results of $n$ comparisons, we can compute the number of wins $w = \sum_{k=1}^n x_k$. With this Bernoulli process outcomes, the likelihood can be defined as \cite{sivia2006data}:
            \begin{equation}
                L(p | \mathbf{x}) \propto p^w (1-p)^{n-w}.
            \end{equation}
            In Bayesian probability theory, for certain likelihood functions, there exist a conjugate prior, where the prior and posterior are in the same family of distributions. This enables fast and analytical computation of the posterior. For the likelihood above, the conjugate prior is known to be a Beta distribution with two shape parameters $\alpha > 0$ and $\beta  > 0$. The posterior Beta density $\pi(p|\mathbf{x}, \alpha_{init}, \beta_{init})$ simply uses the following rule for updates~\cite{fink1997compendium}: 
            \begin{align}
                \alpha &\gets \alpha_{init} + w,\\
                \beta &\gets \beta_{init} + (n-w).
            \end{align}
            With priors of $\alpha_{init}=1$ and $\beta_{init}=1$, we get a uniform prior, as in we do not have any prior preference between items at the beginning of the process of CJ. Henceforth, for notational simplicity, we remove $\mathbf{x}$, $\alpha_{init}$ and $\beta_{init}$ from the equations. As we collect data, the density changes its shape through the updates in $\alpha$ and $\beta$; an example is given in Figure~\ref{fig:ex-prior-post}. Clearly, this update can be done as a sequential process or all together at the end of the data collection, and it can be rapidly performed for a pair for any amount of data. 

            With this framework, we define the probability that $i$ is preferred over $j$, i.e. a different interpretation of probability of winning in \eqref{eq:btm_prob}, as:
            \begin{equation}
                P(i \succ j) = P\left(\pi(p_{[i,j]})> 0.5\right) = 1 - \mathcal{F}(0.5), 
            \end{equation}
            where $\mathcal{F}(\cdot)$ is the cumulative distribution function (CDF) for the Beta PDF $\pi(p_{[j,i]})$.
            Using symmetry, we can calculate the probability that $j$ will be preferred over $i$ as:
            \begin{equation}
            \label{eq:prob_sym}
                P(j\succ i) = 1 - P(i\succ j).
            \end{equation}

            We now expand this analyses for $N$ items and discuss the computation of the distribution over ranks based on this model.

            \subsubsection{Distribution Over the Rank of an Item}

            For a set of $N$ items, we therefore define a $N\times N$ matrix $\mathcal{P}$, where each cell holds a PDF $\mathcal{P}_{[i,j]} = \pi(p_{[i,j]}) ~|~ i\neq j$ defined by a respective $\alpha_{[i,j]}$ and $\beta_{[i,j]}$ updated in a Bayesian manner based on observed data. The diagonal of this matrix is essentially empty, as it does not make sense to construct a preference density for the same item paired with itself. Now, because of symmetry discussed in \eqref{eq:prob_sym}, we are only required to consider the upper triangle of this matrix for updates, which is fast to compute, even for large $N$.

            The $i$th row $\mathcal{P}_{[i,:]}$ captures the relationship between $i$ and other components in the set $I$. Now, to compute the probability that an item is ranked at the top, we must consider all the constituent probabilities that the item dominates each of the other individual items. 
            To be precise, it must simultaneously dominate all other items in the set of all items; hence, this aggregation should be done with the product rule assuming independence between the preferences for $i$th item when compared with each of the other unique items. We can write down the expression for computing this probability as follows (with $1$ being the top rank):
                \begin{equation}
                    P(r_i = 1) = \prod_{j\in I\setminus\{i\}} P(i \succ j).
                \end{equation}
                
                Similarly, we can compute the probability of an item ranked at the bottom as:
                    \begin{align}
                        P(r_i = N) = \prod_{j\in I\setminus\{i\}} P(j \succ i).
                    \end{align}

                For generalisation, specifically for intermediary ranks, for an arbitrary rank $a$, first consider a set $O = I \setminus \{i\}$ with cardinality $|O| = N-1$. Now, for $i$ to be in rank $a$, there must be $a-1$ \textit{dominant} items. From set O, we can pick $z_{a} = C_{N-1,a-1} = \frac{(N-1)!}{(N-a)! (a-1)!}$ combinations without repetitions that can be considered as dominating $i$th item. For every $k$th combination, we then split $O$ into two sets: one for dominant items $D_k$ and the other for dominated items $E_k$, where $|D_k| = a-1$, and $|D_k| + |E_k| = |O|$. For $k$th combination with $D_k$ and $E_k$, the component probability that $i$ is ranked $a$ is: 
                \begin{equation}
                    P(r_i = a | D_k, E_k) = \prod_{s\in D_k} P(s \succ i) \prod_{t\in E_k} P(i\succ t).
                \end{equation}

                Expanding on this, the total probability that $i$ is ranked $a$ can be expressed as:
                \begin{equation}
                    \label{eq:prob_rank}
                    P(r_i = a) = \sum_{k=1}^{z_{a}} P(r_i = a | D_k, E_k),
                \end{equation}
                which for a range of $a\in [1,N]\subset\mathbb{N}$ is a discrete probability distribution, and adheres to the property $\sum_a P(r_i = a) = 1$.
                The expected  (i.e. average or the first moment) rank of an item $i$ can thus be computed using \cite{feller1968stirling}:
                \begin{equation}
                \label{eq:exp_bcj}
                    \mathbb{E}[r_i] = \sum_{a} a P(r_i = a).
                \end{equation}
                
                Now, the number of component combinations that construct the complete probability density for an item is $\sum_{l=1}^N z_l$. Thus, to repeat the procedure for all items, it would require $N \sum_{l=1}^N z_l$ components to be identified and computed. For example, with $25$ items, there will be over $419$m components. While each component is fast to compute, with a large number of components it may be computationally expensive to compute the complete probability density for all items. 

                A straightforward way to combat this expense of computing the expected rank of an item in \eqref{eq:exp_bcj} is to use a form of numerical integration. In fact, a simple Monte Carlo (MC) integration~\cite{mackay1998introduction} with a large \textit{enough} number of samples would be effective in this case (as we illustrate in the next section). To perform MC estimation of the expected rank of an item $i$, we first take samples from the respective row of the matrix $\mathcal{P}$: this generates a sample vector $\mathbf{x}'_i = (x_{[i,j]}')_{j \in [1, N] \wedge i \neq j }^\top$, where $x_{[i,j]}' = \lfloor X \rceil ~|~ X \sim \mathcal{P}_{[i,j]}$. This allows us to count the number of times $i$ has won a comparison $w' = \sum_{j \in [1, N] \wedge i \neq j } x_{[i,j]}'$. Naturally, the rank is $r_i' = (N + 1) - w'$; c.f. with \eqref{eq:btm_r}. For $R$ samples, we can then estimate the expected rank of $i$ as follows:
                \begin{equation}
                \label{eq:mc_prob_rank}
                    \mathbb{E}[r_i] = \frac{1}{R} \sum_{k=1}^R r_i'[k],
                \end{equation}
                where $r_i'[k]$ is the $k$th sampled rank for $i$. 
                
                The standard error of this estimate is known to be $\frac{\sigma_s}{\sqrt{R}}$ with $\sigma_s$ as the standard deviation of the samples~\cite{koehler2009assessment}. In other words, the standard error reduces at the rate of $\frac{1}{\sqrt{R}}$. It is typical to use $10$k samples for this approximation method. So, in this case, we would need $10000N$ samples to estimates ranks for all items, which can be done efficiently in a standard desktop computer, even for large $N$.

                To determine the final rank of the items, we sort items by their the expected ranks:
                \begin{equation}
                \label{eq:bcj_r}
                   r_{i\in I} = (N + 1) - \text{argsort}(\mathbb{E}[\mathbf{r}]).
                \end{equation}

                We present an illustrative synthetic example in the following section.

        \subsubsection{An Illustration}
        \label{sec:illus}
\begin{figure}[t!]
                \centering
                \includegraphics[width=7.5cm]{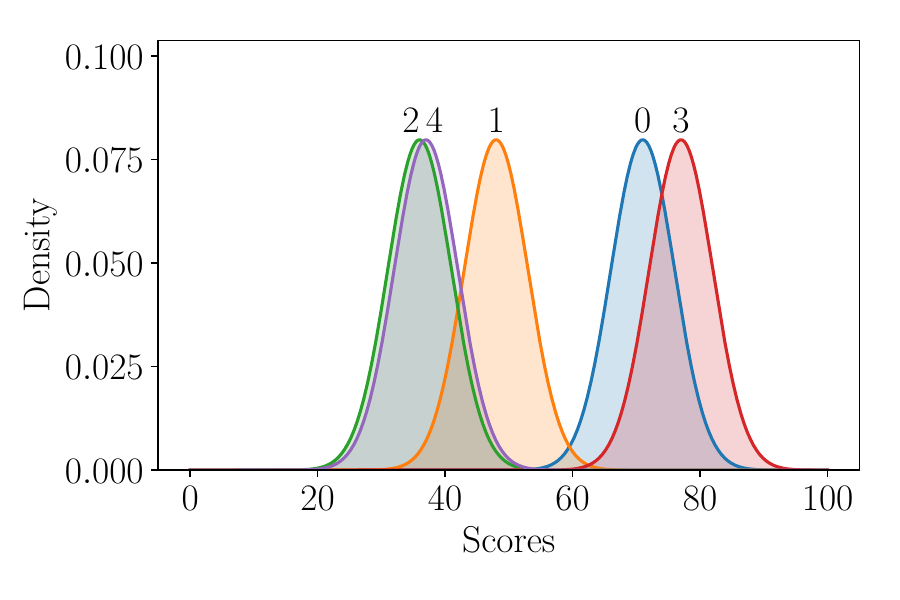}
                \caption{An illustration of five items with Normally distributed scores. Here, the mean vector for the items is $\bm{\mu} = (71, 48, 36, 77, 37)^\top$ and $\sigma = 5$ to represent uncertainty around the mean scores. A simulated paired comparison entails sampling from a pair of these distributions, and whichever yields the higher score wins.}
                \label{fig:samplingdistribution}
            \end{figure}

            We consider a set of \textit{five} items with respective scores and the associated uncertainties, as shown in Figure~\ref{fig:samplingdistribution}. We assume that the scores are Normally distributed (as per Thurstone's original work). To generate the means of these distributions, we uniformly sampled $N$ numbers between $30$ and $90$. Typically, it is often acceptable to have $\pm 10$ score difference between markers when the scores are on a scale between [0,100]. So, we set the two standard deviations of the distributions to $10$, i.e. $2\sigma = 10$. It should be noted that these assumptions about score ranges and standard deviations are solely for illustration purposes. The method presented in this paper is not reliant on these, and it can work with arbitrary distributions over the scores.

            With Normal distributions over scores, we can compute the probability distributions over ranks for any item using the formula in \eqref{eq:prob_rank} as we can calculate the probability that $i$ dominates $j$ as follows \cite{hughes2001evolutionary}:
            \begin{equation}
                \label{eq:prob_norm}
                P(i \succ j) = \frac{1}{2}\left[1 + \text{erf}\left(\frac{m}{\sqrt{2}}\right)\right],
            \end{equation}
            with $m = \frac{\mu_i - \mu_j}{\sqrt{\sigma_i^2 + \sigma_j^2}}$ where $\mu_i$ and $\mu_j$ and means of the Normal distributions for $i$ and $j$, and the associated standard deviations are $\sigma_i$ and $\sigma_j$. The function $\text{erf}(\cdot)$ represents the Gauss error function \cite{andrews1998special}.

            \begin{figure*}[t]
                        \centering
                        \includegraphics[width=1.75\columnwidth]{"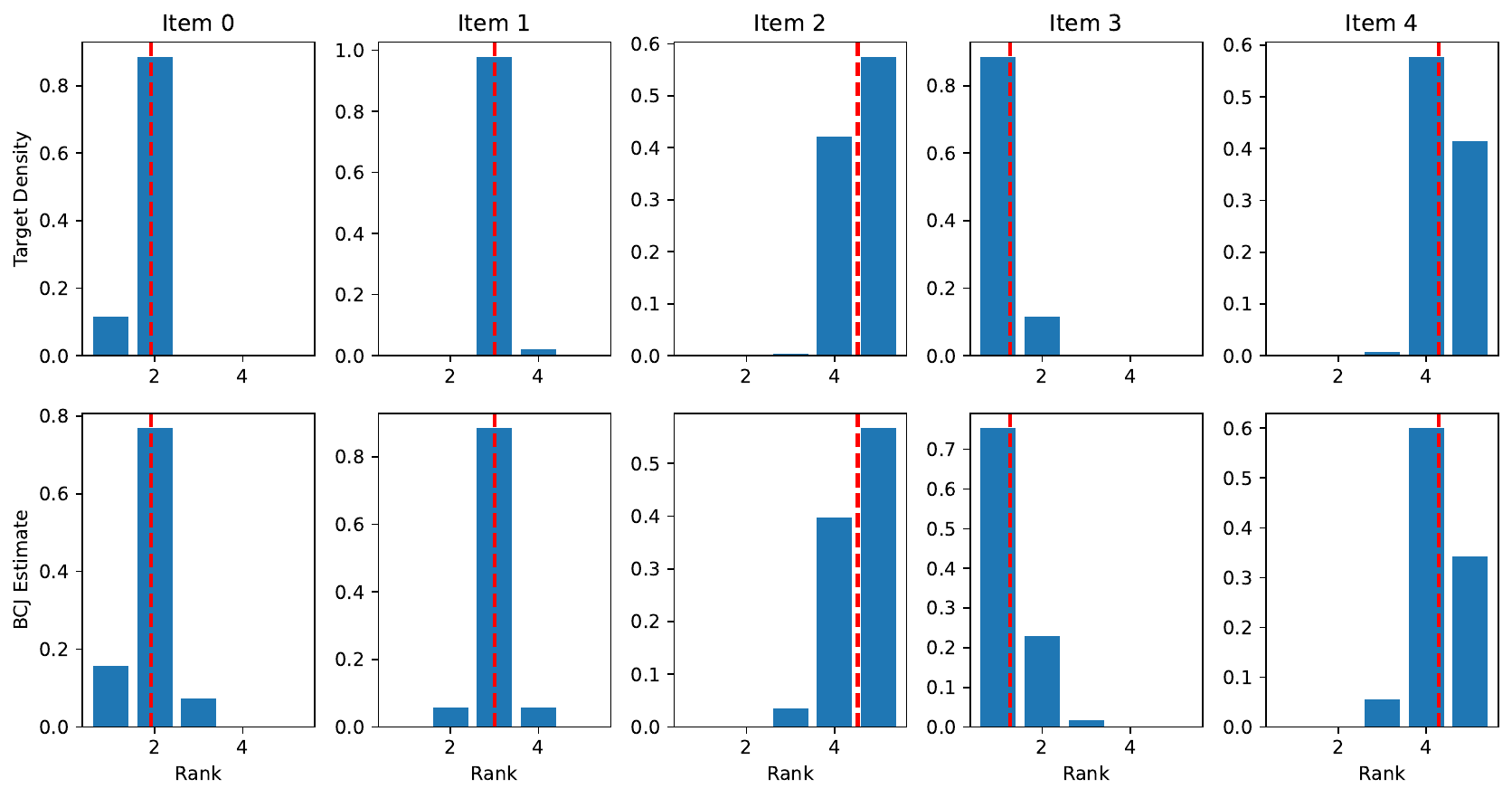"} %
                        \caption{Probability distributions of ranks of items presented in Figure~\ref{fig:samplingdistribution}. The top row shows the densities calculated directly from the Normal distributions over the scores using \eqref{eq:prob_rank}. The bottom row shows the estimated rank distributions using our proposed BCJ method after $N\times K = 5 \times 10 = 50$ pairwise comparisons (driven by our entropy based active learning method presented in Section~\ref{sec:al_ent}). The red dashed vertical line in each panel depicts the expected rank for relevant density. Clearly, our method can well-approximate the target densities, as well as the expected ranks $\mathbb{E}[\mathbf{r}]$. }
                        \label{fig:actual_vs_predicted}
            \end{figure*}

            \begin{figure}[t]%
                    \centering
                    \includegraphics[width=7.25cm]{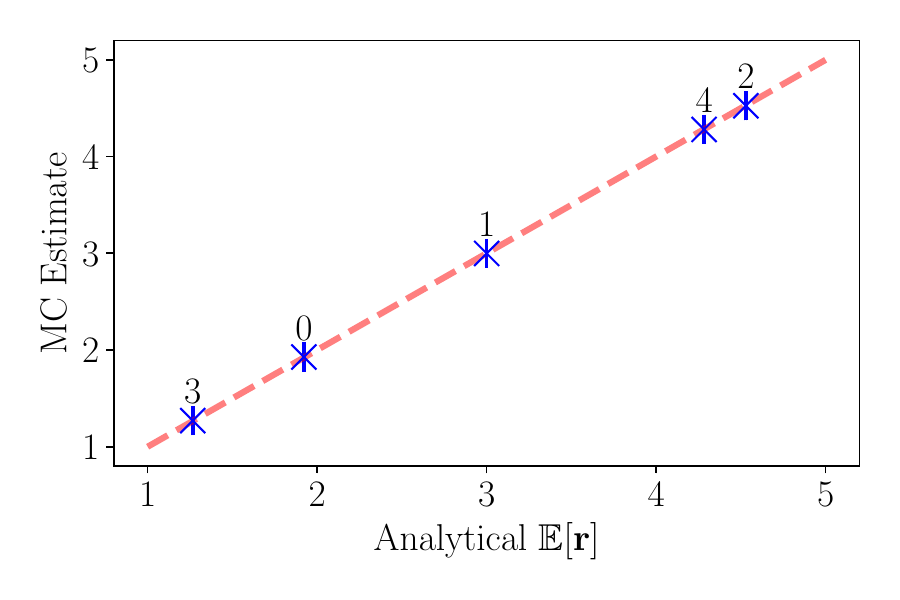}
                    
                    \caption{Comparison between analytical in \eqref{eq:prob_rank} and Monte Carlo estimates (with $10$k samples) in \eqref{eq:mc_prob_rank} of the expected rank of items $\mathbb{E}[\mathbf{r}]$ for our proposed BCJ method after $N\times K = 5 \times 10 = 50$ comparisons as in Figure~\ref{fig:actual_vs_predicted}. The crosses show the mean MC estimate, and the vertical error bars represent the respective uncertainty in approximation, and as expected they are reasonably small for the $10$k samples. The red dashed line shows when there is perfect agreement between the analytical and estimated values, and we see that the average MC estimates are (almost) perfect.}
                    \label{fig:mc_bcj_expected_values}%
                \end{figure}

            In Figure~\ref{fig:actual_vs_predicted}, we show the target distribution over ranks for the items in Figure~\ref{fig:samplingdistribution}, computed using \eqref{eq:prob_rank} and \eqref{eq:prob_norm}. In this case, to emulate the result of a comparison, we sample from the pair of densities and whichever produces a higher score wins the duel. After completing $N\times K = 5 \times 10 = 50$ comparisons using our proposed BCJ method, we can well-approximate the target distributions. To measure how close the estimated distribution is, we use the Jensen-Shannon divergence (JSD). This measure is based on the Kullback–Leibler divergence, with some notable differences, including that it is symmetric, and it always has a finite value between $0$ and $1$~\cite{thiagarajan2022jensen} with values $0$ representing a perfect match. In this case, we get JSD values of $0.0299, 0.0254, 0.008, 0.0185,$ and $0.0125$, which are reasonably close to $0$. 

            It should be noted that with the traditional BTM based CJ, we cannot get an estimate of the probability densities over the ranks, and hence, it is impossible to compute an average rank in this manner. In that method, the scores are instead used to rank items. To compare our approach with BTM based CJ, we will therefore use the BCJ expected ranks to identify the ranks of items.

            In Figure~\ref{fig:mc_bcj_expected_values}, we show a comparison between analytical and MC estimates of rank distributions of items with the BCJ process. Clearly, the MC estimates are highly reliable. So, for large $N$, we recommend using MC estimates for generating expected ranks. In this paper, we used the analytical approach henceforward.

            In the next section, we discuss the selection of pairs to evaluate problem and relevant solutions, including our entropy driven approach. 

    \section{Selecting a Pair of Items to Compare}
        \label{sec:pair_method}
        
        One of the key questions when deploying a CJ approach for marking is how do we select the next pair to evaluate (step~\ref{alg:pick_pair} in Algorithm~\ref{alg:opt}) for identifying comparative preference. There are many ways to generate these, see, for example~\cite{jones_davies_2022}, but these are typically \textit{ad hoc} in nature. Also, Ofqual has stated that if the number of pairs becomes too big over the optimum number, then the final ranking becomes less effective, but knowing this optimal number of comparisons is unknown~\cite{ofqual2017}. While CJ is typically fast and offers a good means of ranking items of work, it does little to give any insights into how the model generated its results. 
        
        Our goal in this paper is to provide further insights into the process to the assessors, particularly focusing on the uncertainties as illustrated in the previous section. More importantly, we want to drive the selection of the pairs to be evaluated using the knowledge that we have already gathered and thus facilitate decision-making in an informed manner to potentially reduce the need for many evaluations.

        It should be noted that the traditional stopping criterion is usually until we exhaust a budget on the number of pairs evaluated: here, we assume that the budget is $N\times K$ where $K$ is the multiplier that is often set to $10$~\cite{jones_davies_2022}.
        
        In this section, we describe three ways to identify the next pair to be compared: through complete randomness, using NRP and our novel method using entropy. 
        
        \subsection{Random}
            The random approach uses a method where every pair presented to the user is picked uniformly at random until the budget is reached. This can cause real-world issues with the change that the same pair can be presented to the user, but that would be unlikely, especially as $N$ increases in size. This is effectively a random search method, that is known to be effective for high-dimensional problems~\cite{bergstra2012random}. 
            We use this widely used method~\cite{jones_davies_2022, benton2018comparative} as a baseline for comparison. 
            
        \subsection{No Repeating Pairs}
            This is another approach used within current approaches: it is essentially a round-robin approach, where no repeating pairs occur until we have selected all possible pairs~\cite{jones_davies_2022, ofqual2017}. This ensures that all $N$ items are seen the same number of times, but what item is compared against what item is decided uniformly as random. This prevents the same pairs from being presented to a user until every other pairs have been rated. However, as we have no indication of uncertainty, certain pairs may be selected despite the difference between them being clear. 

        \subsection{Active Learning with Entropy} 
        \label{sec:al_ent}
            We have also developed a novel approach to selecting pairs in the context of CJ, which uses a Bayesian active learning (AL) approach. AL is a subcategory of machine learning in which a learning algorithm can request input or labels from a user or any other information source to label new data points~\cite{settles2009active, knijnenburg2015evaluating, das2016incorporating}. In Bayesian AL, we use a Bayesian model to make predictions, and then actively select the next data points that should be labelled via an acquisition function that identifies the utility of augmenting the dataset with this new data point; see, for instance~\cite{mackay1992information}. This way we collect data efficiently and learn a good model with fewer data points. 

            There are many variants of AL. In this paper, we focus on so-called ``pool-based learning''~\cite{zhan2021comparative} where we have a finite set of options, and we are going to choose one to show to the labeller. The simplest acquisition function in this context is known as \textit{uncertainty sampling}, where the option with the highest uncertainty is selected for labelling~\cite{lewis1995sequential}. 
            
            To be precise, in our context, we have a finite set of pairs of items, and we will select the one with the highest posterior uncertainty. This uncertainty can be measured with \textit{entropy} where higher uncertainty being represented by higher entropy, and for the posterior Beta density of BCJ, it can be computed as \cite{lazo1978entropy}:
             \begin{multline}
                 H\left[\pi(p_{[i,j]})\right] = \ln B(\alpha,\beta) - (\alpha - 1) \psi(\alpha) \\- (\beta - 1) \psi(\beta) + (\alpha+\beta-2)\psi(\alpha+\beta)
             \end{multline}

             In this paper, we propose to locate the cell in the matrix $\mathcal{P}$ that has the highest entropy and select that pair to be presented to the assessor for making a choice on the preferred item.  

                \begin{figure}%
                    \centering
                    \subfloat[Entropy score for each unique combination after every pairing round. 
                    A higher entropy value shown in lighter colour depicts higher uncertainty.
                    \label{subplot:ent_vis}]{{\includegraphics[width=\columnwidth, trim={10mm 12mm 10mm 25mm}, clip=true]{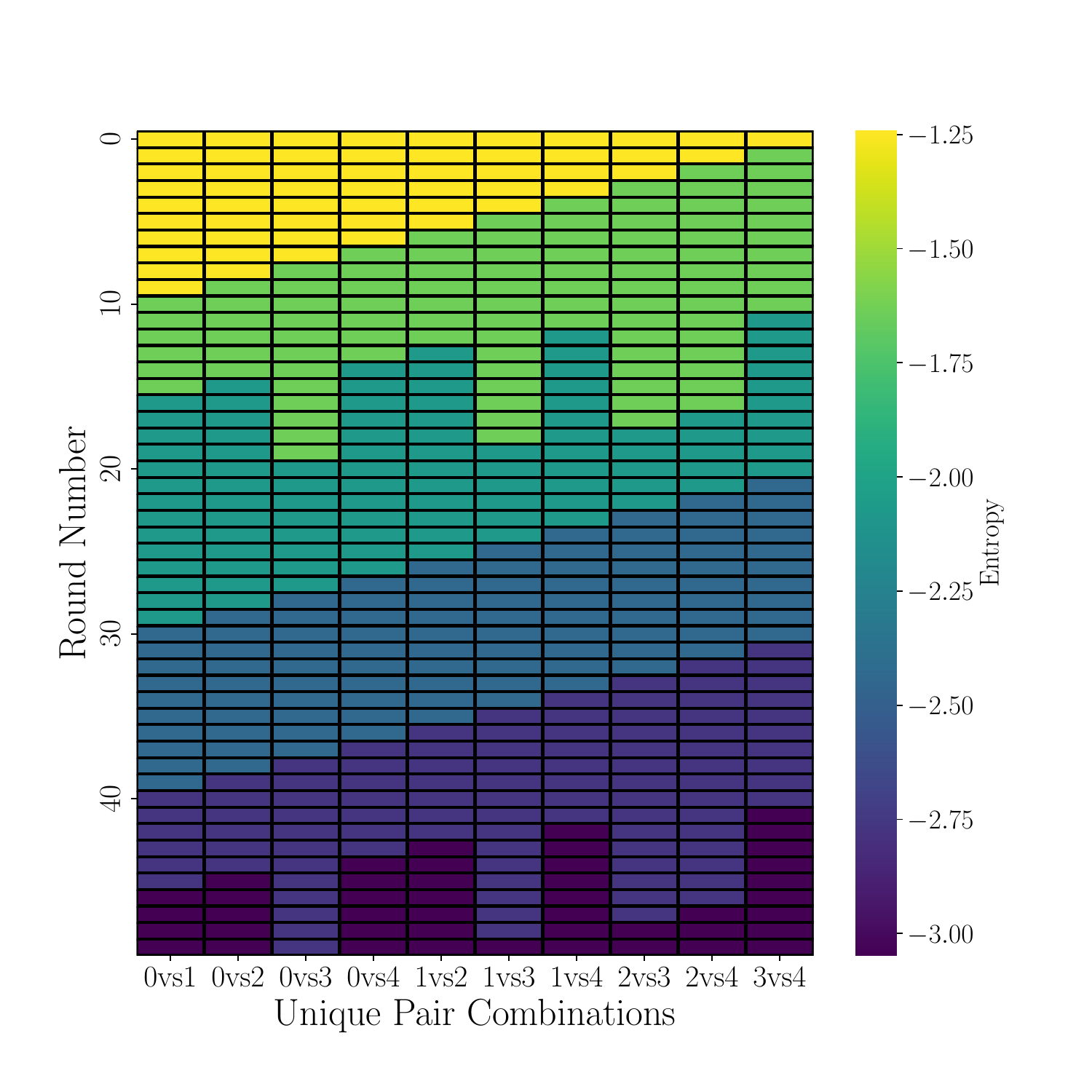} }}%
                    \qquad
                    \subfloat[Progression of highest entropy value after every AL round. \label{subplot:ent_line}]{{\includegraphics[width=\columnwidth, trim={5mm 5mm 5mm 5mm}, clip=true]{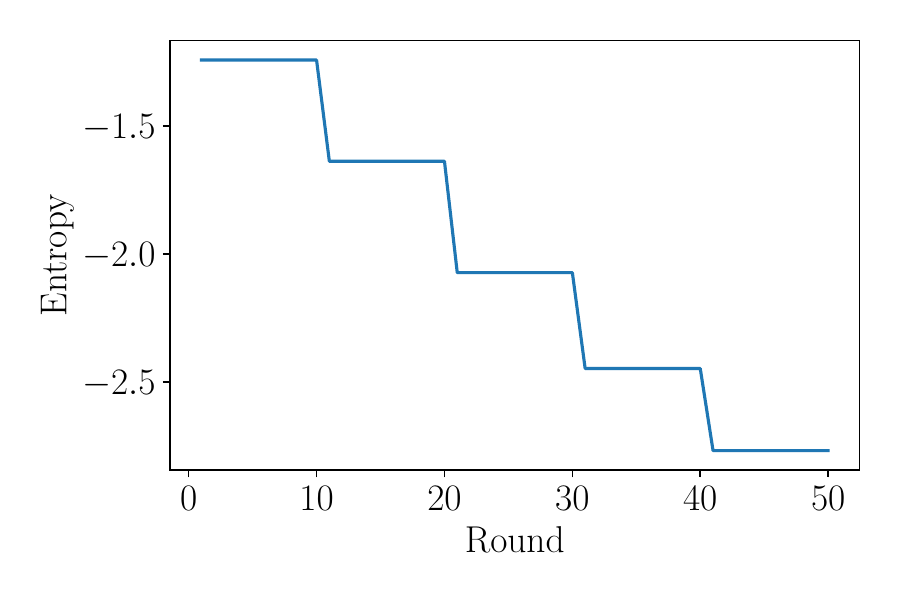} }}%
                    \caption{Illustration of uncertainty sampling using entropy (\textit{top}) for the five items in Figure \ref{fig:samplingdistribution} after $N\times K = 50$ comparisons, and the respective gradual reduction in maximum entropy (\textit{bottom}). As a pair is selected, its uncertainty reduces immediately after data is gathered about preference. The downward trajectory in maximum entropy shows that the model is becoming more accurate over iterations. 
                    }
                    \label{fig:entropy_picking}%
                \end{figure}        
                
                In Figure~\ref{fig:entropy_picking}, we demonstrate the entropy score after each round of comparisons, and the associated selection process. 
                The process involves the algorithm calculating the entropy value for each pair combination in $\mathcal{P}$ to see which pair has the highest value, and then selecting that pair to be presented. However, if there are multiple combinations at the same entropy score, 
                the algorithm will randomly select a pair of values from the list of combinations with the same entropy value. This process will repeat until the required number of rounds is reached. As we can see, the process may be similar to a round-robin approach, but our method would adapt to the changing uncertainties in the target densities in Figure~\ref{fig:samplingdistribution}. 

    \section{Experiments and Discussions}
        \label{sec:experiment_discussion}

        In this section, we will state our findings, analyse them and discuss what we believe they represent and mean. 

        In our reading of the literature, we found that the suggested budget for the number of comparisons were $N\times K = 10N$ \cite{jones_davies_2022}. However, in practice a larger budget is often used. To identify what level of $K$ allows different CJ methods to produce reasonable performance, we ran a range of experiments with $K \in \{5, 10, 20, 30\}$. 
        
        \sloppy As discussed thus far, we have two rank generation methods: BTM and BCJ, and three pair selection methods:  random (R), no repeating pairs (NR), and entropy (E) driven AL. Taking all possible combinations of rank generation and pair selection methods, we can construct a set of \textit{six} approaches for CJ: $S=\{BTM^R, BTM^{NR}, BTM^E, BCJ^R, BCJ^{NR}, BCJ^E \}$. We run $50$ repeated experiments for each approach in $S$ for a given $N$ and $K$, each time starting from scratch, to identify the best. These experiments were conducted with synthetically generated target distributions (following the methods elaborated in Section~\ref{sec:illus}); these were paired, and therefore, we performed Wilcoxon Rank-Sum test on the final results with Bon-Ferroni correction for multiple comparisons \cite{rupert2012simultaneous} at a significance level of $\alpha=0.05$.
        
        Measuring performance of the methods is not straightforward. We consider that targets of scores of items has uncertainty, and they are Normally distributed. Traditional CJ only generates a single rank for items without any uncertainty. To compare results, we use the target distributions to derive the expected rank of each item, and then sorting items by expected ranks gives us a target rank; see Equation \eqref{eq:bcj_r}. This allows us to measure performance via normalised Kendall's $\tau$ rank distance, which measures the difference between two ranking lists. The metric is calculated by counting the discrepancies between the two lists. The greater the distance, the more disparate the lists~\cite{kendall1938new, fagin2003comparing}. The normalised distance ranges from 0 (indicating perfect agreement between the two lists) to 1 (indicating complete disagreement between the lists). For example, a distance of $0.03$ means that only $3\%$ of the pairs differ in ordering. In this paper, when a method progressed, we noted the $\tau$ distance after each paired comparison, and this showed how well the relevant method converged to the target rank. 
        
        It should be noted that BCJ can estimate the whole distribution. So, we can compute JSD, as discussed in Section~\ref{sec:illus}, to identify the agreement between target and BCJ estimated densities.

        In the following sections, we first discuss the performance of different methods in terms of $\tau$ distance. Then we discuss how well BCJ does in estimating the complete target distributions in terms JSD. Finally, we propose yet another method for assigning grade letters to individual items based on the complete probability distribution over rank of an item.

            \subsection{Analysing the Winning Method} 
                \begin{figure*}[t]%
                    \centering
                    \subfloat[\centering Random pairs selector with the BTM approach.\label{subfig:random_btm_10}]{{\includegraphics[width=5.25cm]{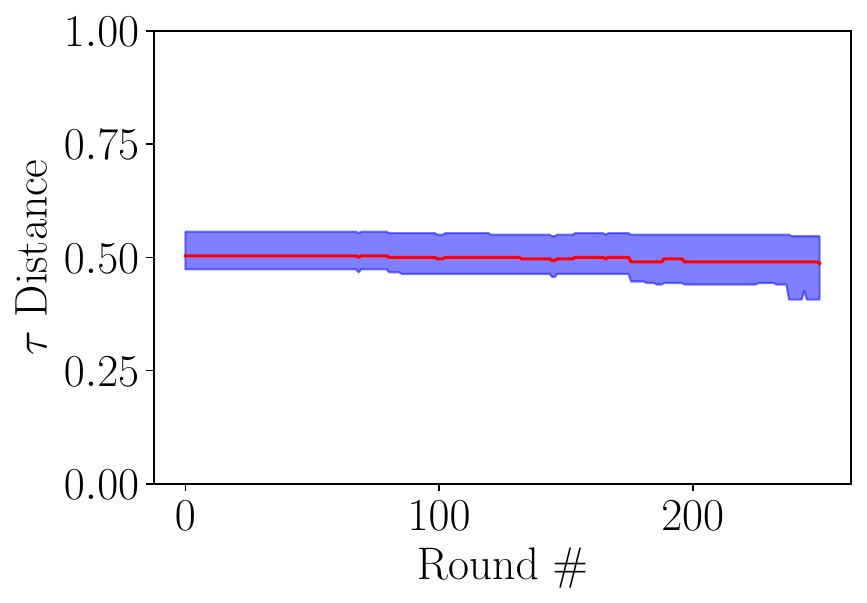} }}%
                    \qquad
                    \subfloat[\centering  No repeating pairs selector with the BTM approach.\label{subfig:nrp_btm_10}]{{\includegraphics[width=5.25cm]{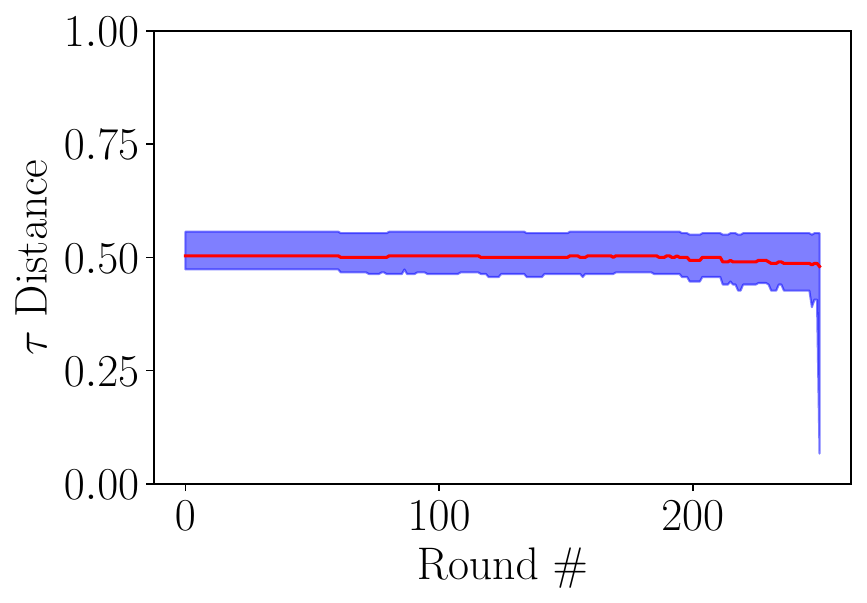} }}%
                    \qquad
                    \subfloat[\centering Entropy pairs selector with the BTM approach.\label{subfig:ent_btm_10}]{{\includegraphics[width=5.25cm]{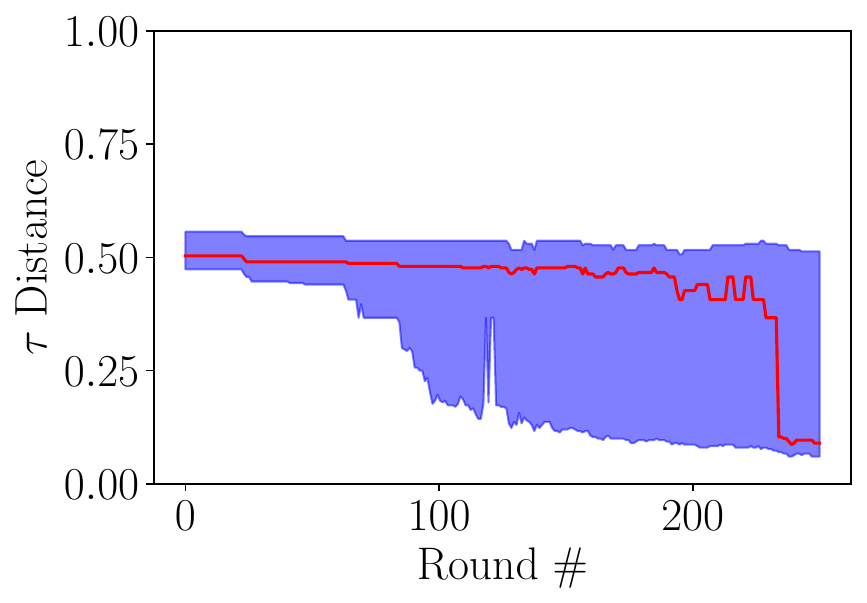} }}%
                    
                    \subfloat[\centering Random pairs selector with the Bayes approach.\label{subfig:random_bayes_10}]{{\includegraphics[width=5.25cm]{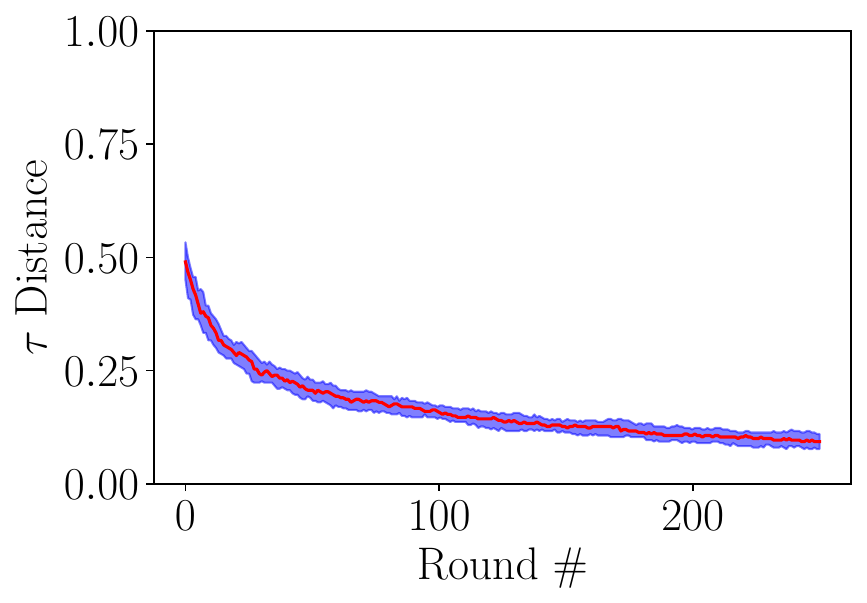} }}%
                    \qquad
                    \subfloat[\centering  No repeating pairs selector with the Bayes approach.\label{subfig:nrp_bayes_10}]{{\includegraphics[width=5.25cm]{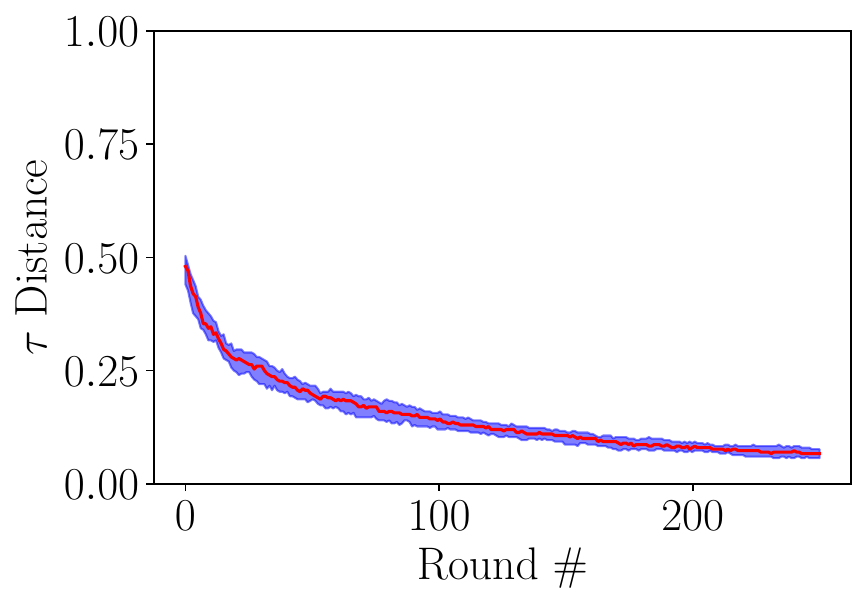} }}%
                    \qquad
                    \subfloat[\centering Entropy pairs selector with the Bayes approach. \label{subfig:ent_bayes_10}]{{\includegraphics[width=5.25cm]{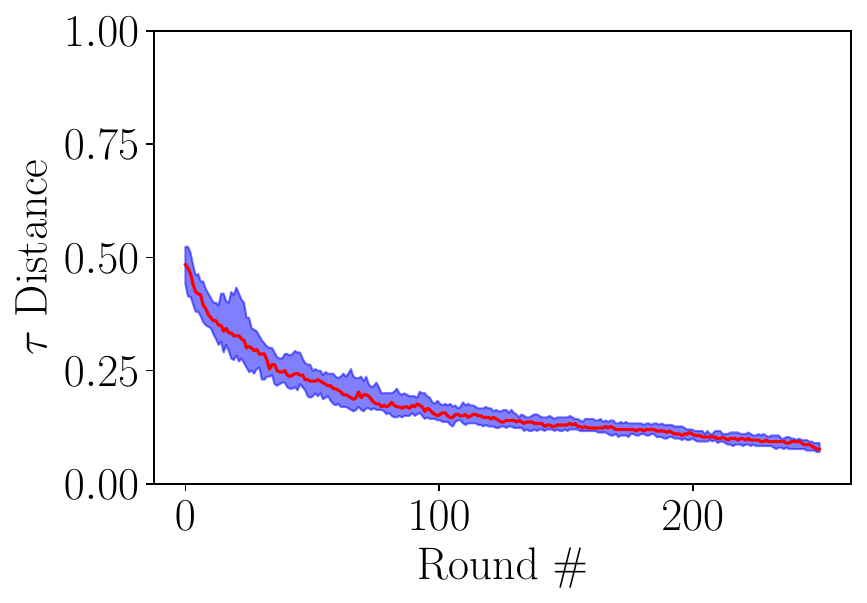} }}%
                    \caption{A comparison of the random (\ref{subfig:random_btm_10}, \ref{subfig:random_bayes_10}), no repeating pairs (\ref{subfig:nrp_btm_10}, \ref{subfig:nrp_bayes_10}) and entropy (\ref{subfig:ent_btm_10}, \ref{subfig:ent_bayes_10}) $\tau$ distance results. The light blue regions show performance between the $25^{th}$ and $75^{th}$ percent quartiles, and the red line depicts the median performance over $50$ repetitions for 25 items where $K=10$, making it a budget of 250 comparisons. The top row shows performances for BTM, while the bottom row shows respective results for our proposed Bayesian approach. Clearly, BCJ outperforms BTM throughout the progress towards the budget. 
                    }%
                    \label{fig:tau_results_n25}%
                \end{figure*}

                In Figure~\ref{fig:tau_results_n25}, we first illustrate the convergence of each CJ approach for $25$ items with a budget of $250$ comparisons. We can see that overall, the BCJ approach has done better in all three pair selection methods. This is consistent across the board, with the BCJ and the novel entropy pair selection method being generally the best combination. Still, the no repeat selection method in combination with BCJ also performs well, but not as well as the combination of our two novel approaches. It is also worth mentioning that the entropy pair selection method positively impacts the BTM CJ approach.

\begin{figure*}[t]%
                    \centering
                    \subfloat[\centering No Repeating pairs selector with the BTM approach. \label{subfig:random_btm_10_k30}]{{\includegraphics[width=7.25cm]{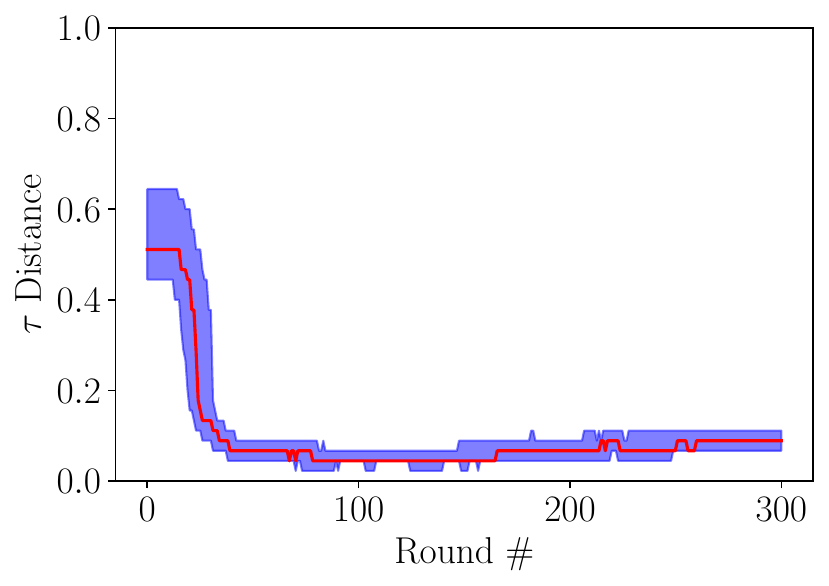} }}%
                    \qquad
                    \subfloat[\centering  Entropy pairs selector with the Bayes approach. \label{subfig:nrp_bayes_10_k30}]{{\includegraphics[width=7.25cm]{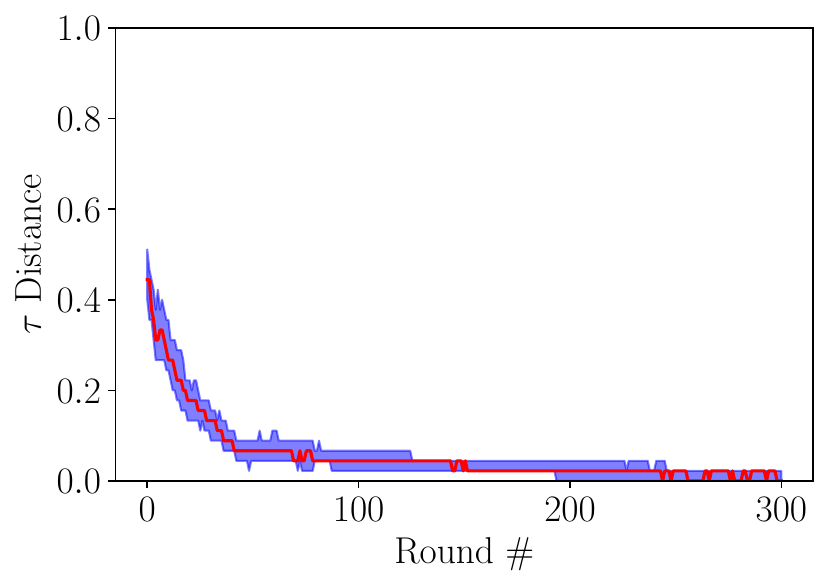} }}%
                    
                    \caption{Convergence plots of the current main method of conducting CJ, a combination of the NR pairing method and BTM (Figure~\ref{subfig:random_btm_10_k30}), and our novel entropy pairing method with BCJ (Figure~\ref{subfig:nrp_bayes_10_k30}). We can see that the BTM method, over time, hits an optimum level but then starts to deteriorate, while the entropy and Bayesian approach always gets more accurate with more data.}

                    \label{fig:tau_results_n10_k30}%
                \end{figure*}

                To investigate Ofqual's claim that the performance of BTM-CJ with no repeating pairs deteriorate with many comparisons~\cite{ofqual2017}, we ran an experiment with $N=10$ and $K=30$ for both the current version of BTM-CJ with no repeating pairs and BCJ with entropy based pair selection. The convergence plots are shown in Figure~\ref{fig:tau_results_n10_k30}. We noted that the performance of BTM-CJ indeed deteriorated over many iterations. However, it is difficult to ascertain the core reasons behind it. We suspect this is because of the uncertainty in determining which one of the pair would be the winner that eventually misleads the BTM algorithm. In contrast, BCJ estimations consistently improved as more data became available. 
                
                \begin{figure*}[h!]%
                    \centering
                    \subfloat[\centering Random pairs selector with the BTM approach.\label{subfig:rpbtm}]
                    {{\includegraphics[width=5.25cm]{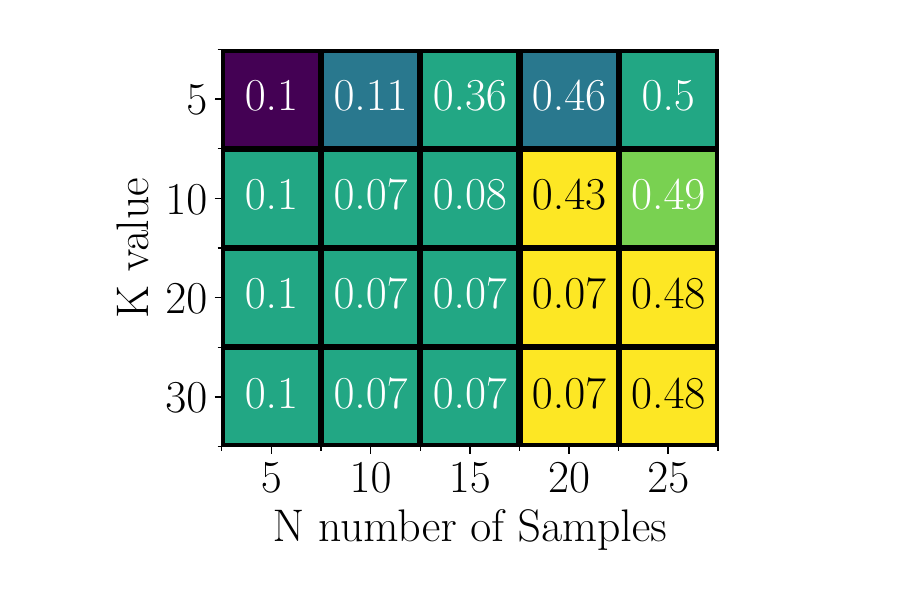} }}
                    \qquad
                    \subfloat[\centering No repeating pairs selector with the BTM approach.\label{subfig:nrpbtm}]{{\includegraphics[width=5.25cm]{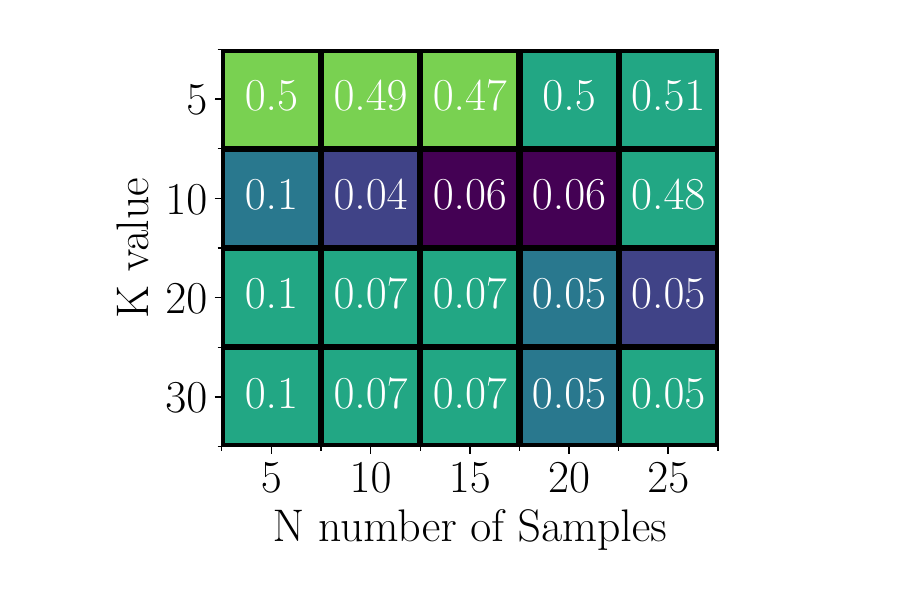} }}%
                    \qquad
                    \subfloat[\centering  Entropy pairs selector with the BTM approach.\label{subfig:entbtm}]{{\includegraphics[width=5.25cm]{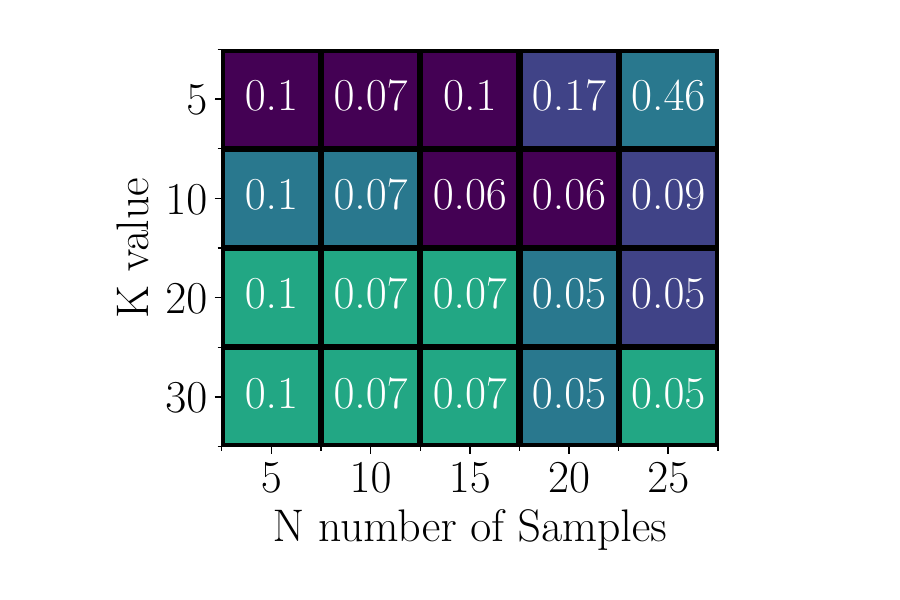} }}%
                    
                    \subfloat[\centering  Random pairs selector with the Bayes approach.\label{subfig:rpbayes}]{{\includegraphics[width=5.25cm]{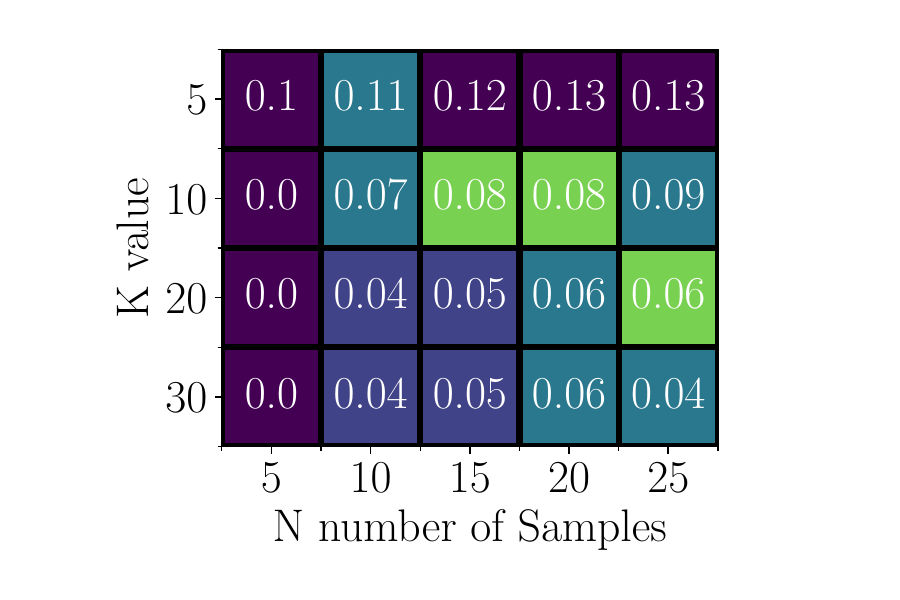} }}%
                    \qquad
                    \subfloat[\centering No repeating pairs selector with the Bayes approach.\label{subfig:nrpbayes}]{{\includegraphics[width=5.25cm]{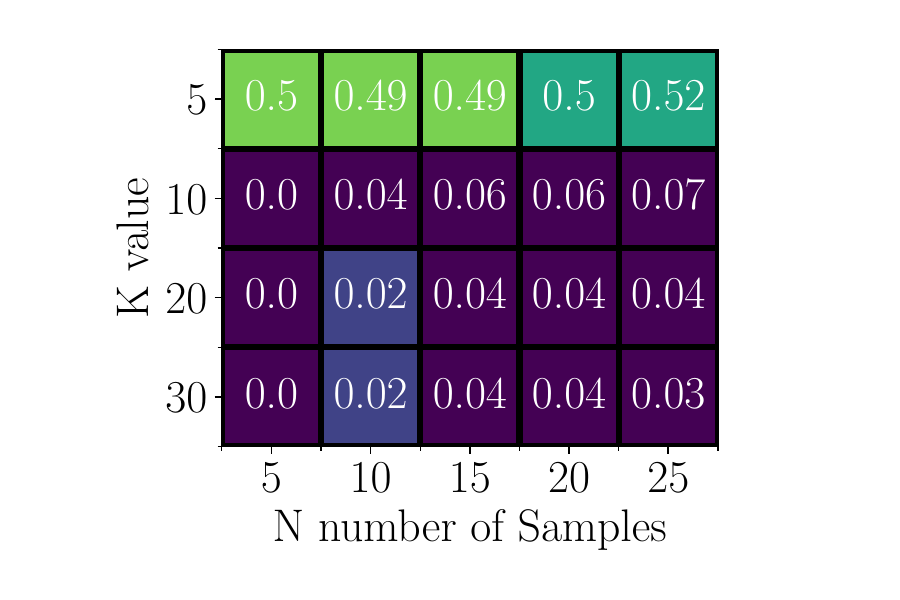} }}%
                    \qquad
                    \subfloat[\centering Entropy pairs selector with the Bayes approach. \label{subfig:entbayes}]{{\includegraphics[width=5.25cm]{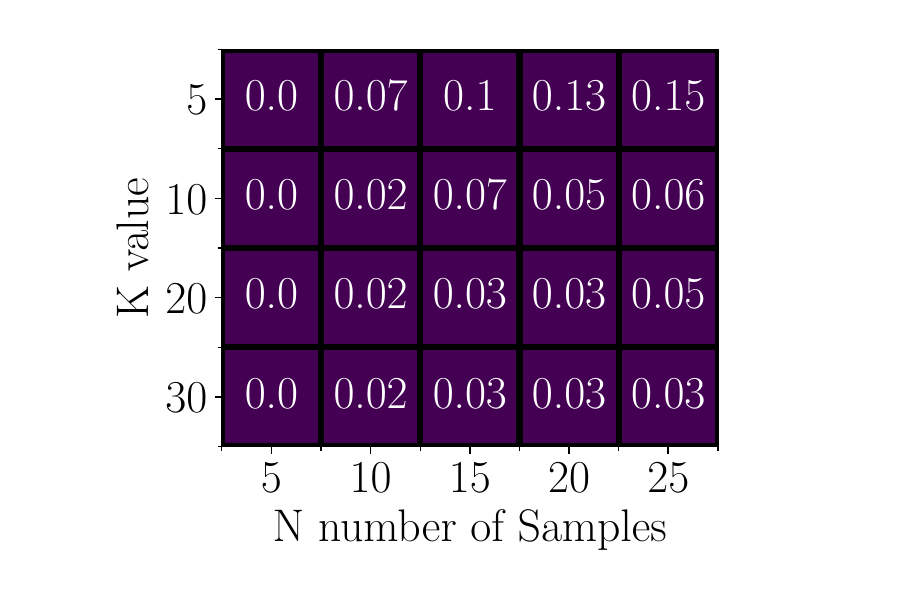} }}%
                    \qquad
                    \subfloat{{\includegraphics[width=8.25cm]{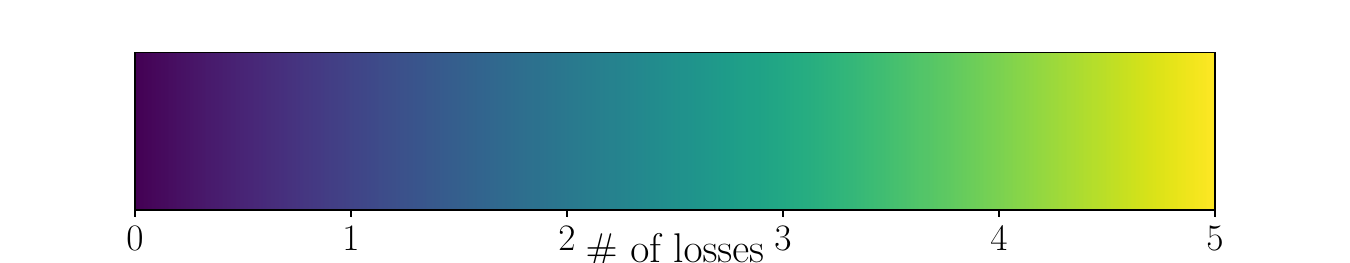} }}%
                    \caption{An illustration of the statistical comparison of results of the random (\ref{subfig:rpbtm}, \ref{subfig:rpbayes}), no repeating pairs (\ref{subfig:nrpbtm}, \ref{subfig:nrpbayes}) and entropy (\ref{subfig:entbtm}, \ref{subfig:entbayes}) selection methods with BTM (\textit{top row}) and Bayesian (\textit{bottom row}) approaches for generating ranks. The plots show the number of times a combination of a ranking method and a pair selection method have been the best, or equivalent to the best, with the darkest colour representing that it was not beaten by any other method for that configuration. The number in white shows the median performance over $50$ repeats for the experimental configuration in the respective cell, with $BCJ^E$ showing the best median performance in $18$ out of the $20$ distinct experiments.}
                    \label{fig:wcrs_results}
                \end{figure*}

                The count of the number of times a method $i$ has been beaten by other methods can be computed with the following expression: $V(i) = \sum_{i\neq j \wedge j\in S} \left[\text{p-value}(i > j) \leq \alpha_{adj}\right]$, where $\text{p-value}(i > j)$ is the binary outcome of comparing $i$ and $j$ with $1$ representing that $i$ has statistically higher value than $j$ (as in $i$ is worse than $j$ in an one-sided manner), and the adjusted significance level is defined as $\alpha_{adj} \gets \frac{\alpha}{m}$, with the original significance level $\alpha = 0.05$ and the number of comparisons $m=5$ for every combination where these tests were performed, using Bon-Ferroni correction for multiple comparisons~\cite{rupert2012simultaneous}.

                These results are shown in Figure~\ref{fig:wcrs_results}. Here, we can see that overall, the Bayesian approaches performed better than BTM. However, the BTM with the entropy-picking method performed reasonably well compared to the other BTM combinations. It should be noted that to use the entropy driven AL with BTM, we must construct Bayesian densities in matrix $\mathcal{P}$.

                In contrast, the Bayes and entropy picking method did considerably better than the rest, with Figure~\ref{subfig:entbayes} showing that this combination was not beaten by any other combination method across all the experiments we conducted. Demonstrating that it is significantly better or, worst case, performed the same as one of the other methods. Interestingly, this shows that our novel approach is better at generating a rank within a lower $K$ value than is suggested; also, the convergence plots in Figures~\ref{fig:tau_results_n25} and \ref{fig:tau_results_n10_k30} support this claim. Additionally, when the $K$ value increases, it still performs well, which is irrelevant to the $N$ value as this doesn't affect its performance.
                
                Therefore, overall we can suggest that the Bayes version as a ranking method has done better, but the combination of Bayes and Entropy has done the best overall. Especially when comparing the current state-of-the-art approach (Figure~\ref{subfig:nrpbtm}) and our two novel approaches (Figure~\ref{subfig:entbayes}).

                We note that in a real-world scenario, in the absence of the information regarding target densities and expected ideal ranks, we cannot compute $\tau$ distances. In this case, we recommend using Figure~\ref{fig:ex-prior-post} for investigating the current state of the preference PDF between any pair of items, and deriving the resulting rank distribution in Figure~\ref{fig:actual_vs_predicted} (bottom row). One can also track the entropy reductions using Figure~\ref{fig:entropy_picking}.

            \subsection{Efficacy in Rank Distribution Predictions} 
                \begin{figure*}[H!]%
                    \centering
                    \subfloat[\centering Random pairs selector JSD median Results.\label{subfig:rp_jsd}]{{\includegraphics[width=5.25cm]{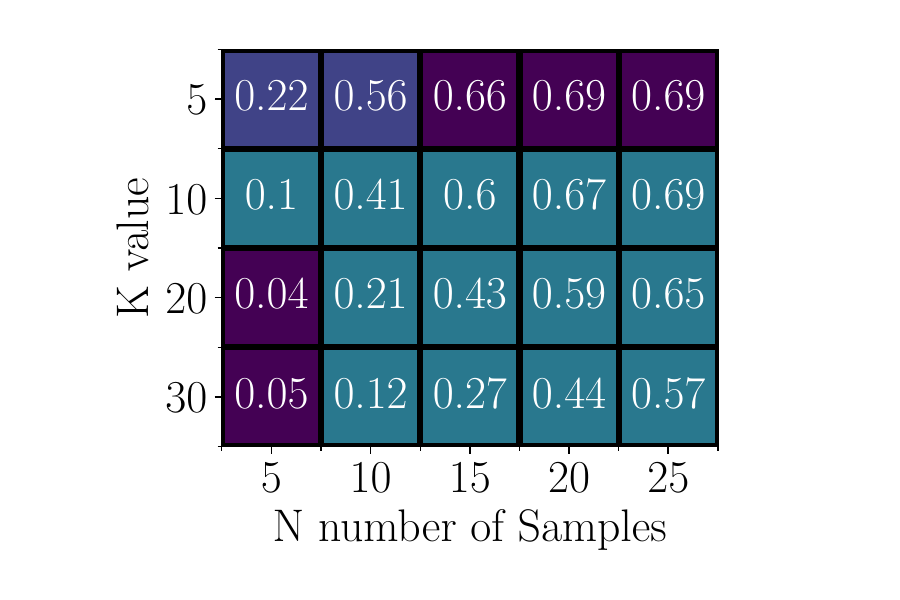} }}%
                    \qquad
                    \subfloat[\centering No repeating pairs selector JSD median Results.\label{subfig:nrp_jsd}]{{\includegraphics[width=5.25cm]{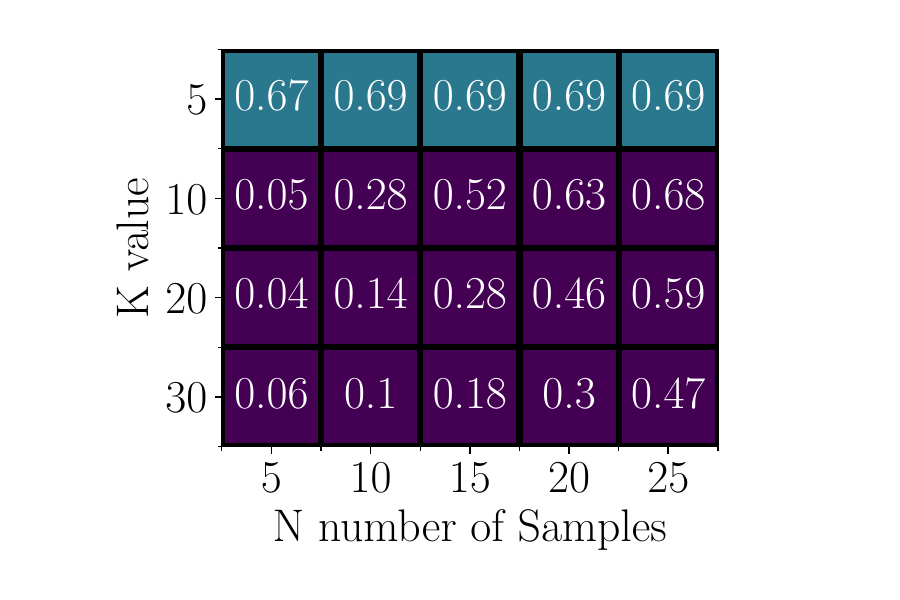} }}%
                    \qquad
                    \subfloat[\centering  Entropy pairs selector JSD median Results.\label{subfig:ent_jsd}]{{\includegraphics[width=5.25cm]{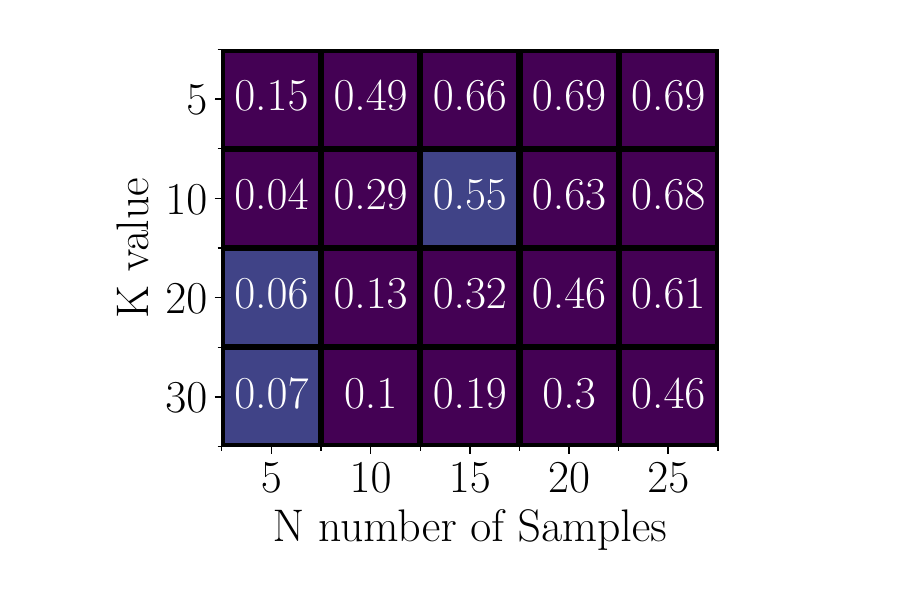} }}%
                    
                    \subfloat{{\includegraphics[width=8.25cm]{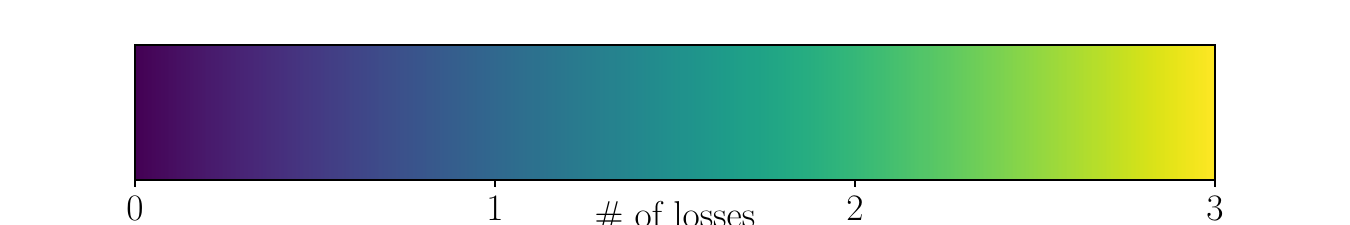}}}%
                    \caption{A comparison of the median JSD results over $50$ repeats of $20$ different experimental configurations for $BCJ^R$ (\textit{left}), $BCJ^{NR}$ (\textit{middle}) and $BCJ^E$ (\textit{right}).}
                    \label{fig:wcrs_results_JSD}
                \end{figure*}

                Due to the BCJ's ability to estimate the complete probability distribution over the rank of an item, we can compare the target densities from the items being compared. Again, in a real-world scenario, this comparison will not be possible, as we do not know the initial target distributions \textit{a priori}.
                
                Here, we used the JSD measure to be able to identify the agreement between our BCJ estimate and actual target distributions. For $N$ items, we deduce $N$ distributions over ranks, and compare with its target counterpart. This comparison gives us $N$ JSD values. We take the worst JSD as reflective of the performance of the current rank distribution, and track this throughout the BCJ process as a measure of progress.

                The results in Figure~\ref{fig:wcrs_results_JSD} show the efficacy of using different pair selection methods when used with BCJ. We see that for $K=5$ using Entropy is the best strategy with random being a close second. Essentially, when there is a lack of data with respect to the number of items being compared, random becomes competitive. However, it seems that no repeating pair strategy is the best for higher $K$ values, with entropy beaten in three instances. While it may be a good strategy with the synthetic targets we constructed, we would still recommend using the proposed uncertainty based approach, i.e. Entropy driven AL, for larger $N$s, as for unknown uncertainty densities over targets, no repeating pairs may not perform as well. 

                Unsurprisingly, comparing Figure~\ref{fig:wcrs_results} and \ref{fig:wcrs_results_JSD}, it is evident that BCJ is better at estimating the expected rank than the complete density of the rank distribution. For example, in Figure~\ref{fig:wcrs_results}, with $N=25$ and $K=30$, $BCJ^E$ has a median $\tau$ distance of $0.03$, which means that only about $3\%$ of all possible pairs, i.e. $9$ out of $300$, differ in order. In contrast, in Figure~\ref{fig:wcrs_results_JSD}, the median of the worst matched item's rank density has a JSD of $0.46$, which is far from the ideal match score of $0$. It is reasonable to expect that with a larger budget on the number of paired comparisons, the rank agreement will improve. 
                
            \subsection{Assigning Grades}
                Different education systems grade assignments differently. For example, in England, exam boards use grades 9 to 1. In contrast, in the educational system in Wales, schools use the more traditional method of $A^*$ to F, while vocational subjects in England and Wales use a Level 2 Distinction$^*$ to Level 1 Pass grading system. Typically, these grades are often assigned based on what \textit{percentile} the work is compared to its peers, and these grades are ultimately what the assessors want to provide to the students. Therefore, it is important to be able to provide a possible grade based on the CJ results to help the assessors.

                \begin{figure*}%
                    \centering
                    \subfloat[\centering Grade Probability distribution presented to the user.\label{fig:proba_single}]{{\includegraphics[width=7.5cm]{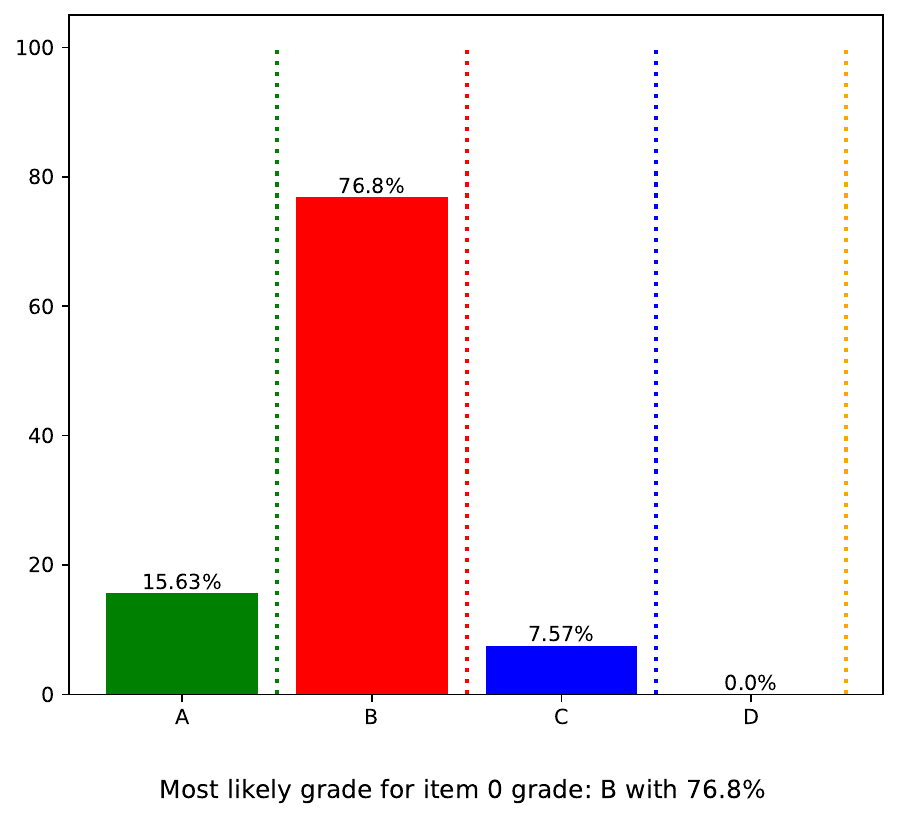} }}%
                    \qquad
                    \subfloat[\centering The cumulative results of the probabilities and the threshold level set to be able to present the expected grade to the user. \label{fig:proba_throshold}]{{\includegraphics[width=7.5cm]{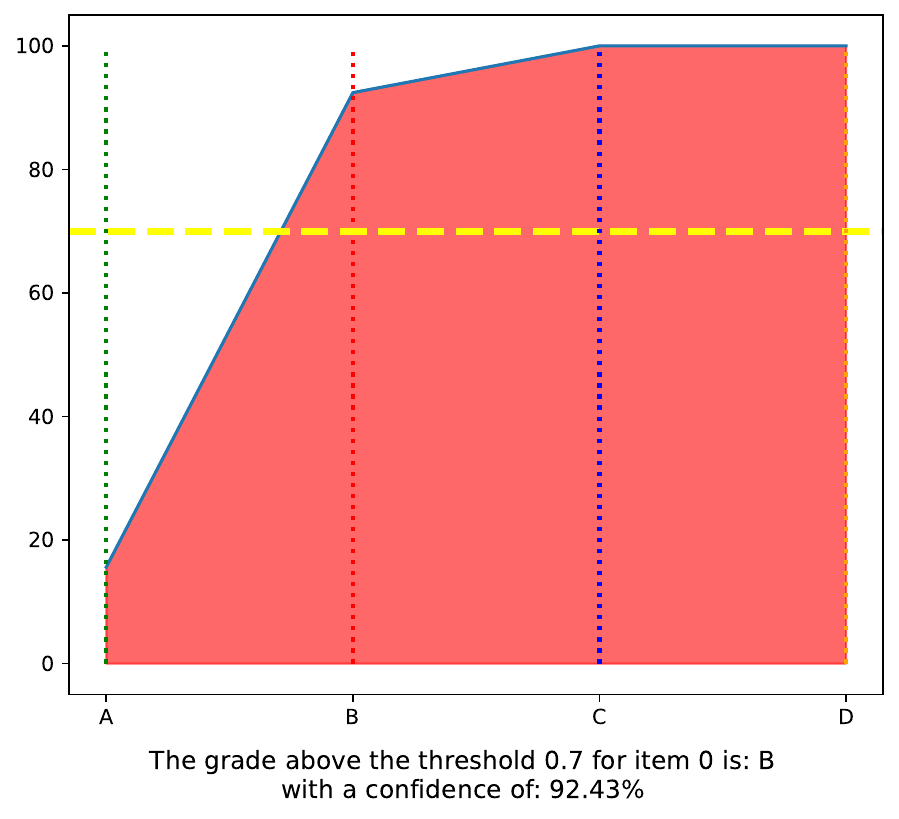} }}%
                    \caption{A figure of the two methods used to present a predicted grade to the user. 
                    The panel on the \textit{left} depicts the probability a student will get a particle grade, while 
                    \ref{fig:proba_throshold} 
                    the panel on the \textit{right} shows the likely grade that meets the threshold level set by the user.}%
                    \label{fig:grade_proba}%
                \end{figure*}
                
                An alternative version of CJ has offered a method of providing a grade within one of their paid subscriptions, which is involved when a nationwide exercise is done, and multiple schools who are part of the service take part~\cite{NMM_accuracy}. However, these are only done twice every academic year and are done by taking a holistic approach from a large number of candidates and using the grading percentages from the previous academic year. 
                
                In this section, we take a different approach that gives greater power to the assessment owner. We propose to use the probability densities over the rank of items to assign a grade to individual pieces of work. Given a discrete probability distribution over the rank of an item, we can compute the probability that an item's rank would be between two values as follows:
                \begin{equation}
                    P(g \leq r_i \leq h) = \sum_{k=g}^h P(r_i = k),
                \end{equation}
                where $g$ and $h$ are the boundary rank of the grade level. Using this we can easily compute the probability that a piece of work lies between a range of ranks, and thus it can be interpreted with the notion of how many pieces of work should get the highest grades, and so on. This determination of grade is then entirely dependent on the assessor's decision on how many students should get what grade; for example, an assessor may decide that only the top $30\%$ would receive a grade $9$ (for an assignment submitted in England).
        
                Figure~\ref{fig:grade_proba} demonstrates this approach through an example of the outcomes after completing the CJ process. The teacher has decided that out of five pieces of work one can receive a grade of A and B, two can receive a C, and one can receive a D. It gives us great insight and therefore presents to the marker, for example, that item $3$ (shown in the \textit{left panel} of Figure~\ref{fig:grade_proba}) has a $15.63\%$ of gaining a grade A, $76.8\%$ a B, $7.57\%$ a C and $0\%$ a grade D. Considering the cumulative probabilities, we can see that there is a $(15.63 + 76.8)\% = 92.43\%$ chance that this item would receive a grade B or above. If the assessor then decides a \textit{threshold of acceptability}, for instance, $90\%$, for achieving a certain grade, we can assign grade B for this work. However, if the threshold was higher, e.g. $95\%$, the work would get a grade of C, as then the cumulative probability would stand at $(15.63 + 76.8 + 7.57)\% = 100\%$ which is greater than the threshold.
        
                The ability to provide predicted grades is only possible due to our BCJ approach, which provides the probability distribution that an item will rank, as seen in Figure~\ref{fig:actual_vs_predicted}. We expect that such probabilistic reasoning renders the assessors greater control over the whole CJ process, with a high level of explainability.
        
        \section{Conclusions}
            \label{sec:conclusion}
            Marking and assessing works of students is an important element of education. However, it takes a long time, can be inconsistent, especially because we are not great at assessing absolute quality. Furthermore, we are starting to see the use of generative AI tools in education and its potential impact on various forms of assessment and associated practices~\cite{dwivedi-et-alchatgpt:2023}.

            However, with the introduction of CJ this has helped alleviate a lot of the quality issues in principle but does come with its own issues. One of the issues is that the paired comparison rank order starts to deteriorate, making the whole model's fit somewhat collapse. However, it is not easy to determine how many comparisons are enough. As the study has shown that the $\tau$ distance score gets worse as the value of $K$ gets bigger. However, 
            The recommended minimum number of comparisons is $N \times 10$, but this study has shown that it struggles after $N \geq 20$, showing that a larger $K$ is required as at the suggested minimum the current CJ with BTM struggles to rank accurately, with results showing that when $N=20$ a $K$ value of $20$ is required to start getting close to the desired rank.
            Nonetheless, our novel BCJ approach does not suffer from this issue, as the more comparisons we make, the more accurate it gets.

            Most importantly, there are issues around using any current form of CJ as a replacement for marking, as the outcome is less transparent~\cite{ofqual2017}. During the design of our new BCJ approach, we focused on addressing the issue of transparency by being able to provide information to the user about how the algorithm has come up with its rank decisions, as well as allowing the user to give input into how it generates the grades as well as giving the information on how it predicted what it has predicted. Therefore, rendering greater transparency compared to the standard approach, and it is computationally affordable too. Future work will look into providing automated feedback based on the ranks predicted with BCJ. 

\section{Acknowledgements}
    \label{}
    Andy Gray 
    is funded by the EPSRC Centre for Doctoral Training in {\emph{Enhancing Human Interactions and Collaborations with Data and Intelligence-Driven Systems}} (EP/S021892/1) at 
    Swansea University. 
    Additionally, the project stakeholder is 
    CDSM and their CTO, Darren Wallace. 
    We would also like to thank Dr 
    Jennifer Pearson 
    for their valuable feedback on the manuscript. 
    For the purpose of Open Access, the author has applied a CC-BY public copyright licence to any Author Accepted Manuscript (AAM) version arising from this submission.
    All underlying data to support the conclusions are provided within this paper.

\appendix
\bibliographystyle{elsarticle-num} 
\balance
\bibliography{cas-refs}

\bio{}
\endbio

\end{document}